\RequirePackage{fix-cm}
\documentclass[twocolumn]{svjour3}          % twocolumn
\smartqed  % flush right qed marks, e.g. at end of proof

\usepackage{array}
\usepackage{bm}
\usepackage{url}
\usepackage{todonotes}
\usepackage{setspace}
\usepackage{natbib}
\usepackage{graphicx}
\usepackage{multirow}
\usepackage{amsmath,amssymb,amsfonts}%
\usepackage{mathrsfs}%
\usepackage[title]{appendix}%
\usepackage{xcolor}%
\usepackage{textcomp}%
\usepackage{manyfoot}%
\usepackage{booktabs}%
\usepackage{algorithm}%
\usepackage{algpseudocode}%
\usepackage{listings}%
\usepackage{colortbl}
\usepackage{comment}
\usepackage{tabularx}

\usepackage[colorlinks=true]{hyperref}
\begin{document}
\sloppy

\title{SMLNet: A SPD Manifold Learning Network for Infrared and Visible Image Fusion}

%\subtitle{Do you have a subtitle?\\ If so, write it here}

%\titlerunning{Short form of title}        % if too long for running head

\author{Huan~Kang      \and
        Hui~Li         \and
        Tianyang~Xu         \and
        Xiao-Jun~Wu         \and    
        Rui~Wang       \and
        Chunyang~Cheng       \and
        Josef~Kittler
}
%\author[1]{\fnm{Chunyang} \sur{Cheng}}\email{chunyang\_cheng@163.com}

%\authorrunning{Short form of author list} % if too long for running head

\institute{H.~Kang, H.~Li, T.~Xu, X. J.~Wu*, R.~Wang and C.~Cheng  \at
              School of Artificial Intelligence and Computer Science\\
              Jiangnan University, Wuxi, 214122, China. \\
              *\email{wu\_xiaojun@jiangnan.edu.cn}           %  \\
%             \emph{Present address:} of F. Author  %  if needed
           % \and
           %    X.~Li \at
           %    College of Computer Science and Technology\\
           %    Zhejiang University, Hangzhou, 310027, China. \\
           %    \email{xilizju@zju.edu.cn}
           \and
              J.~Kitter \at
              Centre for Vision, Speech and Signal Processing\\
              University of Surrey, Guildford, GU2 7XH, UK. \\
              \email{j.kittler@surrey.ac.uk}
}

\date{Received: January, 2025 / Accepted: date}
% The correct dates will be entered by the editor

\maketitle

\begin{abstract}
Euclidean representation learning methods have achieved promising results in image fusion tasks, which can be attributed to their clear advantages in handling with linear space.
However, data collected from a realistic scene usually has a non-Euclidean structure, evaluating the consistency of latent representations from paired views using Euclidean distance raises challenges.
To address this issue, a novel SPD (symmetric positive definite) manifold learning is proposed for multi-modal image fusion, named SMLNet, which extends the image fusion approach from the Euclidean space to the SPD manifolds. Specifically, we encode images according to the Riemannian geometry to exploit their intrinsic statistical correlations, thereby aligning with human visual perception. The SPD matrix fundamentally underpins our network's learning process. Building upon this mathematical foundation, we employ a cross-modal fusion strategy to exploit modality-specific dependencies and augment complementary information. To capture semantic similarity in images' intrinsic space, we further develop an attention module that meticulously processes the cross-modal semantic affinity matrix. Based on this, we design an end-to-end fusion network based on cross-modal manifold learning. Extensive experiments on public datasets demonstrate that our framework exhibits superior performance compared to the current state-of-the-art methods.
Our code will be publicly available at https://github.com/Shaoyun2023.
% \myrevisedcolor{Our code will be publicly available soon.}

% \PACS{PACS code1 \and PACS code2 \and more}
% \subclass{MSC code1 \and MSC code2 \and more}
\end{abstract}

\section{Introduction}\label{sec1}

With the development of multi-modal sensor technology, image fusion is playing an increasingly important role in the field of computer vision. This task integrates image data from different imaging sensors to obtain richer and more accurate information about the scene. % across different modalities. 
Multi-modal image fusion has many applications in various practical fields such as medical imaging ~\citep{1,2}, remote sensing monitoring ~\citep{3,4}, semantic segmentation ~\citep{5}, and object tracking ~\citep{zhu2024unimod1k}, etc.

In the past few decades, many effective image fusion methods have been developed, which can roughly be divided into two categories: image fusion techniques based on traditional methods and image fusion techniques based on deep learning networks.

The introduction of many traditional methods such as multi-scale transformation ~\citep{6} and low-rank representation ~\citep{zhang2021exploring,7,8} has significantly contributed to the image fusion task. However, these methods typically depend on isolated feature representations or basic combinations, such as raw pixel values, local descriptors, or histograms, none of which adequately model global feature relationships. Additionally, they neglect the interactions among local features. To address this, our method introduces the covariance matrix as a mathematical descriptor to quantify statistical dependencies among feature blocks, identifying positive and negative correlations of features. This is essential for uncovering the latent dependencies between global and local features in an image.

Compared to the traditional image processing techniques, deep learning architectures~\citep{9,zhang2023visible,wang2023review,liu2017multi} have shown significant effectiveness in the field of image fusion due to their superior ability in extracting high-level 
image features, and learning of nonlinear transformations. In the current research, methods based on Convolutional Neural Networks (CNNs) ~\citep{10,liu2017multi} have shown to be effective in the field of image fusion. These methods adopt an end-to-end learning paradigm ~\citep{11,deng2024mmdrfuse} that trains the network directly from pairs of input images to produce the fused result, capturing image characteristics at different levels from low-level appearance information to higher-level semantic content. However, CNN extracts image features through convolution operations in the local receptive field, and it underestimates the long-range dependencies in the image, which impedes the image fusion performance. For instance, in multi-focus image fusion ~\citep{liu2020multi,liu2017multi}, maintaining relations between foreground and background usually requires considering contextual information from the entire image scene.

In 2021, the Vision Transformer (ViT) ~\citep{dosovitskiy2020vit} appeared as a revolutionary architecture, shifting the deep learning paradigm from traditional convolution operations to processing images by dividing them into fixed-size tokens, which can capture long-range dependencies between image segments. Subsequent applications, especially in image fusion tasks~\citep{qu2022transmef,ma2022swinfusion,zhao2023cddfuse,li2024crossfuse,cheng2025textfusion}, capitalized on this method's powerful capabilities. Although significant progress has been made, current vision attention models extremely rely on Euclidean algebraic operations, causing the output of attention to neglect the geometric dependencies within high-dimensional data.

% Despite the significant progress, these models overly focus on the self-attention mechanism itself, sacrificing cross-modal interaction processes. Yet, in fusion tasks, such associative information is crucial\cite{zhao2023cddfuse} \cite{xu2023murf}. 
In computer vision, data often exhibits rich intrinsic structures and hidden associations, particularly in multi-modal fusion tasks where samples within a single modality show strong feature-space correlations, while cross-modal relationships remain sparse and underutilized. This gap is effectively bridged by the SPD manifold, composed of symmetric positive definite matrices, which inherently preserves second-order statistical characteristics through its geometric structure. By doing so, it becomes ideal for modeling cross-modal dependencies, where covariance directly quantifies shared variations (\textit{e.g.}, thermal-texture alignment), enabling precise statistical interaction modeling even when inter-modal relationships are weak.

To effectively model intrinsic cross-modal statistical correlations, our approach leverages the geometric properties of SPD manifolds to represent image features, overcoming the limitations of traditional Euclidean methods. By constructing composite covariance matrices that decompose relationships into self-correlation and cross-correlation terms, we explicitly quantify implicit intra-modal and inter-modal dependencies while preserving the manifold's geometric constraints (symmetry, positive definiteness). Unlike Euclidean approaches, this SPD-based paradigm enables statistically rigorous fusion through intrinsic manifold operations, ensuring that the fused feature representations not only capture nonlinear interactions but also align with the true distribution of multi-modal data.

\begin{figure}
    \centering
    \includegraphics[width=1\linewidth]{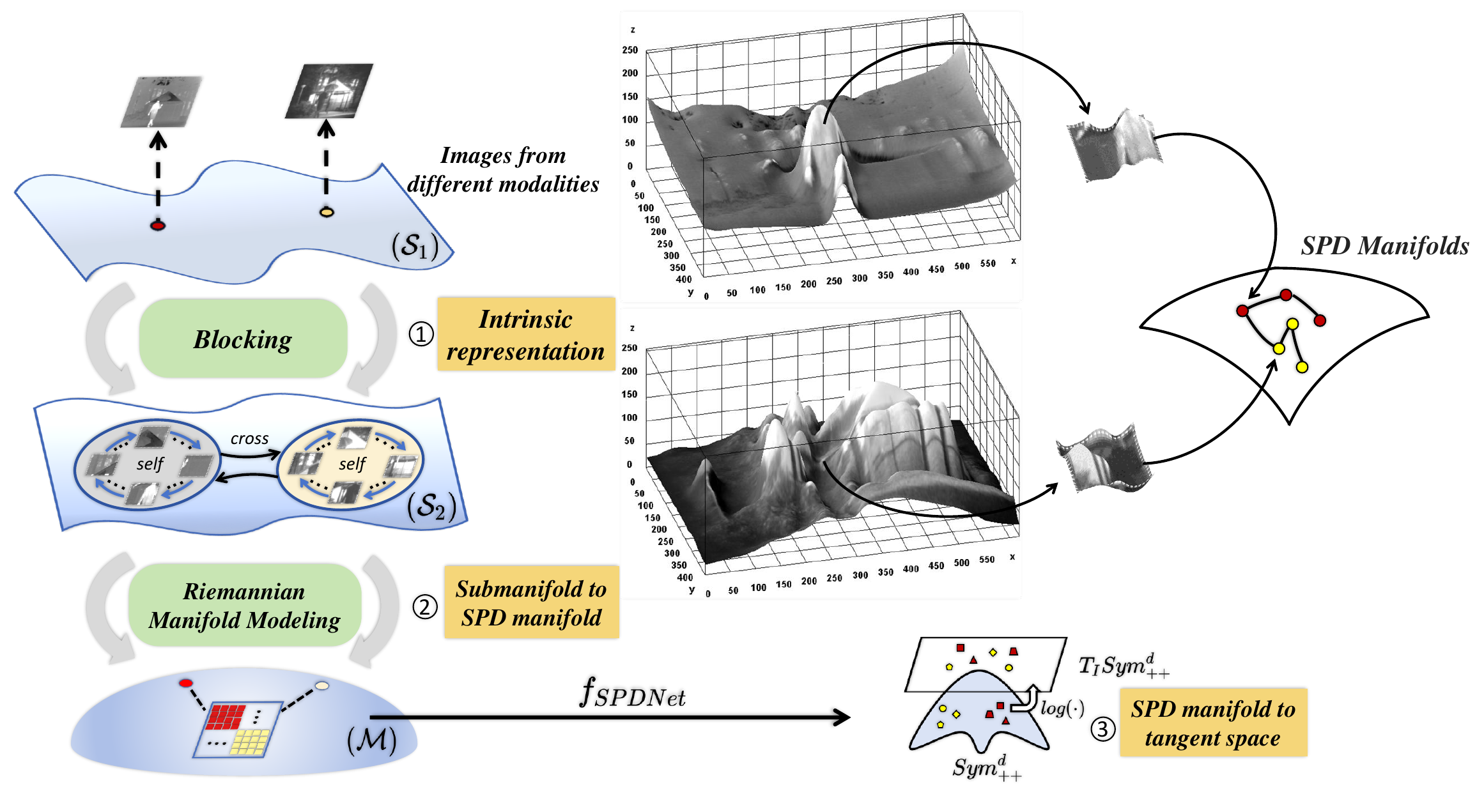}
    % \captionsetup{justification=justified,singlelinecheck=false}
    \caption{A new modeling approach based on the representation of Riemannian manifold. We divide the images from two sources of data into patches, allowing the network to focus on local details and global context information. Where the cross-modal image pairs are considered to be on the manifold \(\mathcal{S}_{1}\), while the manifold \(\mathcal{S}_{2}\) where the patches reside can be regarded as a manifold structure that is homeomorphic to \(\mathcal{S}_{1}\). \(\mathcal{M}\) denotes the SPD Riemannian manifold, \(Sym_{++}^{d}\) represents the symmetric positive definite manifold, and \(T_{I}Sym_{++}^{d}\) is the tangent space of the SPD manifold at the \(d\times d\) identity matrix \(I\).}
    \label{idea}
\end{figure}

Based on this, we propose an infrared and visible image fusion algorithm based on the SPD manifold. This model pioneers the application of Riemannian manifold networks ~\citep{thanwerdas2023n,brooks2019riemannian,nguyen2021geomnet,huang2017riemannian} to multi-modal image fusion tasks, mining the raw pixel characteristics and their statistical relationships to guide the fusion process. This method not only preserves the inherent geometric properties of image data on the Riemannian manifold but also facilitates the interaction of cross-modal information. On one hand, we design an SPD manifold attention module (SPDAM), which, through the introduction of a mixed attention mechanism, strengthens the statistical correlation within each modality and across different modalities. On the other hand, geometric computations on Riemannian manifolds enable a more refined capture and utilization of the spatial nonlinear characteristics, thereby providing a better fit to the intricate geometric structure inherent in image data. Based on this, we model the covariance among image patches, effectively achieving the learning of global correlation relationships, and the main idea of the proposed method is shown in Fig. \ref{idea}.
Our main contributions are as follows:

The contributions of this work can be summarized as follows:
\begin{itemize}
\item For the first time, the Riemannian manifold network is introduced into the image fusion task. The proposed method exhibits unique advantages due to its adaptation to the non-Euclidean distribution of image data and the consideration of its intrinsic geometric structure.
\item A new manifold attention module is designed. Through global correlation representation learning, it strengthens the semantic correlation between features and effectively guides the fusion task.
\item A novel cross-modal fusion strategy has been introduced, which integrates features on the SPD manifold to enhance intrinsic semantic coherence across modalities.
\item Experimental results on public benchmark datasets indicate that our fusion network exhibits superior fusion performance compared to existing fusion methods.

\end{itemize}

\section{Related Work}

\subsection{Covariance Matrix}
The covariance matrix is an important tool for describing the statistical linear dependence between variables in a dataset ~\citep{wang2012covariance}. It offers a comprehensive  perspective on multivariate data in terms of of second-order moment statistics and provides a mathematically rigorous method for quantitative modeling of the interrelationships between variables. 

The potential of the covariance matrix has been effectively utilized in many branches of statistics. However, its application in pattern recognition is relatively underestimated. It has been proven that covariance matrices are effective in various scenarios. For example, in image set classification ~\citep{wang2022learning},~\citep{wang2021symnet}, the covariance matrix can represent the statistical characteristics of a set of images and characterize the image set as a distribution of feature points. This property of covariance can be used to perform more complex tasks, such as recognizing facial expressions or motion patterns. In medical imaging ~\citep{chen2023riemannian}, the covariance matrix is used to capture morphological changes in various tissues or structures, and by studying these changes, disease characteristics can be revealed, helping doctors make more accurate diagnoses.

However, Riemannian networks using covariance matrices are primarily designed for classification tasks, and their extracted features often fall short of meeting the specific demands of image fusion. Moreover, the impact of inter-modal regional correlations on fusion results remains understudied, especially in multi-modal scenarios like multi-spectral satellite imagery and medical image fusion. These correlations can uncover relationships between different modalities, such as scene features from diverse sensors, or help filter irrelevant information.

Considering the advantages of covariance modeling, Our SMLNet innovatively employs a composite covariance matrix architecture to mathematically model both intra-modal feature distributions (\textit{e.g.}, thermal radiation in infrared images and texture details in visible images) and inter-modal joint distribution patterns. This dual statistical modeling mechanism not only completely preserves the unique feature representations of each modality, but more importantly establishes precise statistical dependencies between cross-modal features, thereby achieving efficient suppression of redundant information while enhancing complementary features during the fusion process.

% needed in second column of first page if using \IEEEpubid
%\IEEEpubidadjcol

\subsection{The Basic Layer of SPD Neural Network}
Since traditional machine learning and classification methods operate primarily in Euclidean space, they are ill-suited for processing SPD matrices. These methods fail to account for the intrinsic Riemannian manifold structure of SPD matrices and may distort their geometric properties. Fortunately, researchers have proposed an innovative neural network architecture: SPDNet ~\citep{huang2017riemannian}, which is specifically designed to adapt to the unique features of SPD matrices. In its overall structure, it is similar to traditional Euclidean neural networks, including transformations, nonlinear activations, and the final classification stage. The difference lies in the fact that the SPDNet architecture includes specialized layers to ensure the outputs remain on the manifold, utilizing operations such as Riemannian metric, exponential and logarithmic mappings, as well as geodesic distances compatible with manifold geometry ~\citep{brooks2019riemannian}. The basic layers can be divided into three categories:

BiMap Layer: This layer is designed to reduce the dimension and enhance the discriminative features while preserving the SPD matrix characteristics of the data. By applying a bilinear mapping, the BiMap layer can map SPD matrices from a high-dimensional SPD manifold to another low-dimensional SPD manifold, while maintaining the geometric structure and positive properties of the data. The specific mapping formula is:
\begin{equation}
    X_{k} =f_{b} (X_{k-1};W_{k})=W_{k}X_{k-1}W_{k}^{T}, 
    \label{BiMap}
\end{equation}
where \(X_{k}\)  is the newly mapped SPD matrix and \(X_{k-1}\) is the input SPD matrix to the \(k\)-th layer. \(W_{k}\) is the transformation matrix, which should be row full-rank, and if its columns are orthogonal,  \(W_{k}\) belongs to the compact Stiefel manifold \(St(d_{k},d_{k-1})\), where \(d_{k}\) is the dimension of the mapped SPD matrix after the transformation, typically smaller than \(d_{k-1}\), to achieve dimensionality reduction.

ReEig Layer: It introduces a nonlinear activation function within SPDNet while preserving the SPD matrix structure. The role of this layer is similar to the ReLU activation function, which increases the complexity of decision boundaries and enhances the model's expressive capacity. Similarly, the ReEig layer introduces non-linearity to the Riemannian manifold by rectifying eigenvalues to a lower bound threshold, preventing degradation issues associated with very small eigenvalues, and allowing the model to capture more complex and abstract features. The formula is as follows:
\begin{equation}
    X_{k}=f_{r}^{(k)}(X_{k-1})=U_{k-1}max(\epsilon I,\Sigma_{k-1})U_{k-1}^{T},
    \label{ReEig}
\end{equation}
where the matrices \(U_{k-1}\) and \(\Sigma_{k-1}\) are the results of performing an eigenvalue decomposition (EIG) on \(X_{k-1}\), such that \(X_{k-1}\) is factored as \(U_{k-1}\Sigma_{k-1}U_{k-1}^{T}\), \(\epsilon\) denotes a rectification threshold, \(I\) represents an identity matrix, and \(max(\epsilon I,\sigma_{k-1})\) is obtained by element-wise maximum comparison, with the diagonal entries determined as follows:
\begin{equation}%公式
\begin{aligned}
A(i,i) = 
\begin{cases} 
\sum_{k-1}(i,i) & \text{if } \sum_{k-1}(i,i) > \epsilon \\
\epsilon & \text{if } \sum_{k-1}(i,i) \leq \epsilon 
\end{cases},
\end{aligned}
\label{EIG}
\end{equation}
where \(A\) is a diagonal matrix.

The key to the LogEig layer is that it offers a means to effectively ``straightening" the curved manifold space into a flat Euclidean space while preserving the SPD property (positive-definiteness) by performing logarithmic operations on the eigenvalues obtained from an EIG (eigenvalue decomposition) of the matrix. This makes linear operations feasible in the Euclidean space:

{
\small
\begin{equation}
    X_k = f_l^{(k)} \left( X_{k-1} \right) = \log \left( X_{k-1} \right) = U_{k-1} \log \left( \Sigma_{k-1} \right) U_{k-1}^T,
    \label{LogEig}
\end{equation}
}
where for a SPD matrix \(X_{k-1}\), one first performs the eigenvalue decomposition \(X_{k-1} = U_{k-1} \Sigma_{k-1} U_{k-1}^T\), where \(U_{k-1}\) is an orthogonal matrix comprised of eigenvectors, and \(\Sigma_{k-1}\) is a diagonal matrix of corresponding eigenvalues. Following this, the logarithm of the diagonal elements of \(\Sigma_{k-1}\) is computed to obtain the diagonal matrix of log-eigenvalues \(\log(\Sigma_{k-1})\). Finally, the SPD matrix \(X_k\) is reconstructed through \(U_{k-1} \log(\Sigma_{k-1}) U_{k-1}^T\), thus \(X_k\) represents \(X_{k-1}\) in the log domain.

\begin{figure*}
    \small
    \centering
    \includegraphics[width=1\linewidth]{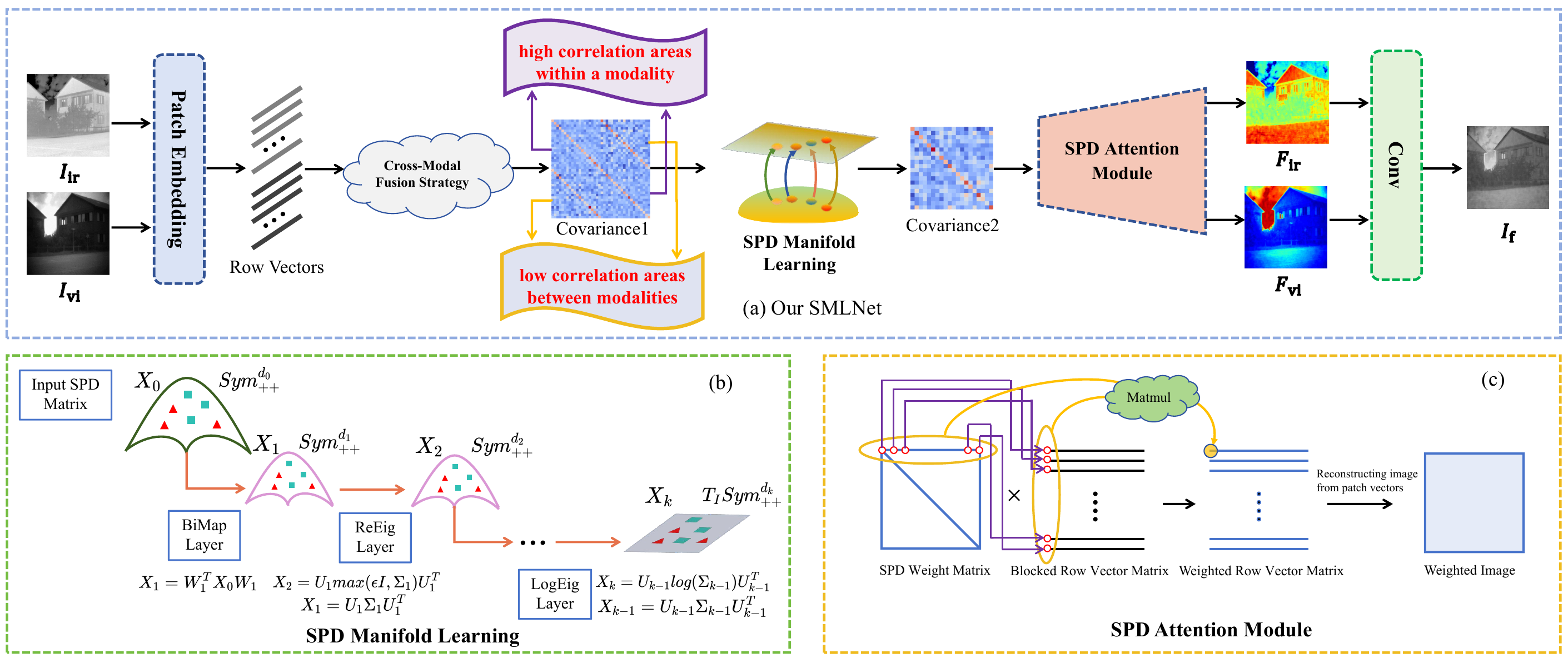}
    % \captionsetup{justification=justified,singlelinecheck=false}
    \caption{The workflow of our SMLNet: (a) The main framework of our fusion network is defined. (b) Details of  manifold learning. (c) The Riemannian attention weighting operation, which is tailored specifically for fusion task. Our approach begins by leveraging the non-Euclidean structure of the data and modeling it in manifold space. The manifold network aggregates cross-modal information while the SPD Attention Module (SPDAM) dynamically allocates attention weights to adaptively align cross-modal image features. The final output is generated through a weighted fusion of the two modal convolutional components.}
    \label{Framework}
\end{figure*}

\subsection{Deep Learning Tasks Based on Attention Mechanism}
In deep learning and computer vision, attention mechanisms aim to emulate the human ability to focus selectively on relevant visual information. While traditional CNNs process all input regions uniformly, such an approach can be suboptimal for complex or cluttered data, see ~\citep{wang2018non,zhang2019self}. With the development of large-scale classification tasks and other vision tasks, researchers began to explore how to make CNNs focus more on important local information in the image to learn discriminative feature representations. This pursuit has led to the development of two principal attention paradigms: spatial attention and channel attention, which are widely used in image processing tasks. Specifically, spatial attention focuses on identifying the importance of specific spatial regions within an image, whereas channel attention operates by emphasizing critical feature channels.

Hu et al. proposed a structural unit called SEBlock (Squeeze-and-Excitation block) ~\citep{hu2018squeeze} based on the inter-relationships between channels. It dynamically calibrates channel feature responses by explicitly modeling the dependencies between channels. However, the SEBlock's approach is limited in that it uses global average pooling to calculate channel attention, which may not be suitable for inferring fine-grained channel attention, and it does not consider the equally important spatial attention, thus failing to achieve semantic-level perception of the context of objects in a scene.

CBAM (Convolutional Block Attention Module) ~\citep{woo2018cbam} introduces spatial attention on top of this. It is a lightweight attention module that refines the input feature maps by sequentially applying channel and spatial attention mappings. Unlike channel attention, spatial attention focuses on the ``where" question in the input feature maps, identifying which areas of the image parts are information-rich, in order to emphasize or suppress these areas. This is usually achieved by processing each spatial location of the feature maps to generate a spatial attention map that indicates the importance of each location.

 Traditional Euclidean attention mechanisms, fundamentally constrained by their reliance on linear dot-product operations, fail to capture intrinsic statistical correlations in data with complex manifold structures. We overcome this limitation by generalizing attention computation to SPD manifolds, explicitly modeling higher-order statistical dependencies via covariance interactions across modalities. Through Riemannian manifold optimization, SMLNet ensures preservation of multi-modal geometric priors while capturing nonlinear feature covariances, thereby constructing geometry-aware cross-modal representations.

\section{The Proposed Method}
In this section, we provide a detailed description of the SMLNet network framework, the cross-modal SPD fusion strategy, the manifold learning network and the 
SPD attention module (SPDAM), as well as an explanation of our training process and loss function.

\subsection{The Network Framework}
The proposed fusion network (SMLNet) architecture is
shown in Fig. \ref{Framework}(a), the source images are first encoded into overlapping patch vectors, and then the covariance of the encoded patch row vector matrix is computed via the ``Cross-Modal Fusion Strategy" to detect statistical relationships between different patches. Through covariance computation, we not only capture the feature relationships within a single modality but also use algorithms to highlight low-correlation yet information-rich features across modalities. (This part will be explained in detail in Section \ref{cross}) In this way, the statistical representations within and between modalities are not only based on the original image space but also incorporate deep statistical information across multiple modalities, providing a rich data basis for manifold learning. Furthermore, to ensure the non-Euclidean geometric structure of the image, our manifold learning module completes cross-modal attention interaction within the Riemannian space, and the learned weight coefficients are ``flattened" back to Euclidean space through logarithmic mapping. Specifically, we designed a Riemannian attention weighting module based on the principle of convolution operations, injecting manifold statistical relevance into the image. Finally, the weighted image features are subjected to deep feature extraction through a series of convolutional layers to obtain the final fusion result.

The source images \(I_{\text{ir}}\) and \(I_{\text{vi}}\) are the infrared image and visible light image, respectively. To acquire a global spatial information representation, we first segment each image into patches with overlapping regions. For each patch, we arrange all the pixels in a sequential order to form a row vector. For each image modality, a matrix of row vectors corresponding to the rearranged image block matrices can be obtained. Specifically, let \(P_{\text{ir}}\) and \(P_{\text{vi}}\) be the patch sets of the infrared and visible light images,with \(p_{\text{ir}_{i}}\) being the \(i\)-th patch of \(I_{\text{ir}}\), and \(p_{\text{vi}_{i}}\) being the \(i\)-th patch of \(I_{\text{vi}}\). For each \(p_{\text{ir}_{i}}\) in \(P_{\text{ir}}\), we can construct the row vector \(x_{\text{ir}_{i}}\). And for each \(p_{\text{vi}_{i}}\) in \(P_{\text{vi}}\), we can also construct the row vector \(x_{\text{vi}_{i}}\). Eventually, we obtain the matrix of row vectors \(\mathbf{M}_{\text{ir}}\) for the infrared image and \(\mathbf{M}_{\text{vi}}\) for the visible light image, which can be expressed as:
\begin{equation}
    \mathbf{M}_{\text{ir}} = \begin{bmatrix}
    x_{\text{ir}_1} \\
    x_{\text{ir}_2} \\
    \vdots \\
    x_{\text{ir}_n} 
    \end{bmatrix}, \quad
    \mathbf{M}_{\text{vi}} = \begin{bmatrix}
    x_{\text{vi}_1} \\
    x_{\text{vi}_2} \\
    \vdots \\
    x_{\text{vi}_n} 
    \end{bmatrix},
    \label{Patch}
\end{equation}
where \(n\) denotes the total number of patches.

Subsequently, we computed the covariance matrix ``Covariance1" between the vectorized patches, which encapsulates the correlation representation information of different regions in the image. To ensure that the covariance matrix satisfies the basic properties of the SPD manifold, we perform an SVD on this symmetric semi-definite matrix, and add a small positive perturbation term \(\epsilon\) to the eigenvalue matrix, to ensure that all the eigenvalues are greater than zero:
\begin{equation}
Q=U(S+\epsilon*tr(S))V^{T},
\label{Epsilon}
\end{equation}
where \(\epsilon\) is a very small positive number, usually set as 0.001. \(U\) and \(V^{T}\) represent the orthogonal matrices obtained from the SVD of the original positive semi-definite matrix, respectively. \(S\) represents the diagonal matrix obtained by singular value decomposition, and the resultant matrix \(Q\) belongs to an SPD manifold \(X_{0}\) in the Riemannian space. Then, \(X_{0}\) goes through the SPD manifold network to learn the deep statistical information of different modalities, and embeds the learned feature manifold into the Euclidean space for attention weighting:
\begin{equation}
    F_{\text{mod}} = \text{MatMul}(X_{k}, \phi_{\text{row}}), \quad \text{mod} \in \{ \text{ir}, \text{vi} \}
\end{equation}
where \(X_{k}\) represents the weight matrix, which can also be represented as ``Covariance2". \(F_{\text{mod}}\) represents the final feature map obtained, and \(\phi_{\text{row}}\) represents the matrix of row vectors.

\begin{figure}
    \centering
    \includegraphics[width=1\linewidth]{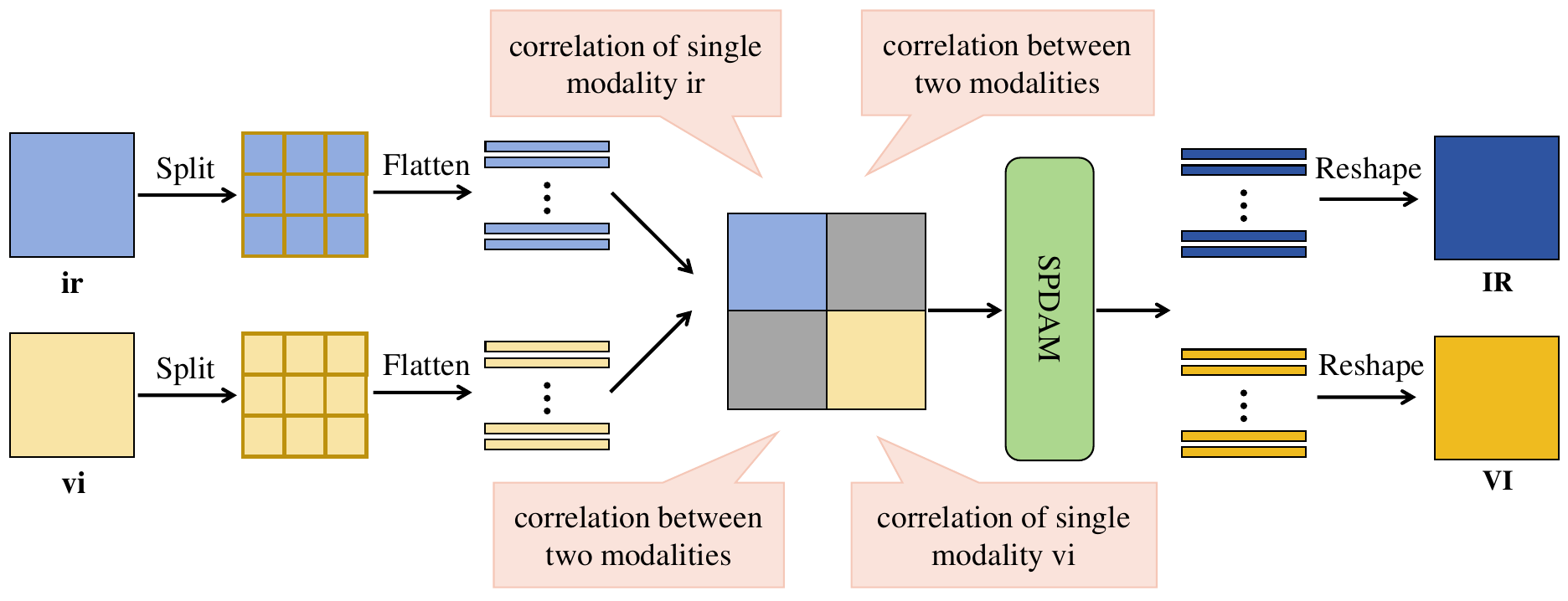}
    % \captionsetup{justification=justified,singlelinecheck=false}
    \caption{The of our fusion strategy. The vectorized image patches are modeled as covariance matrices that represent spatial relationship attributes. Attention weights are adaptively allocated through manifold learning to promote information fusion within a single modality and between modalities.}
    \label{Cross fusion strategy}
\end{figure}

\begin{figure*}
    \centering
    \small
    % \captionsetup{justification=justified,singlelinecheck=false}
    
    \includegraphics[width=1\linewidth]{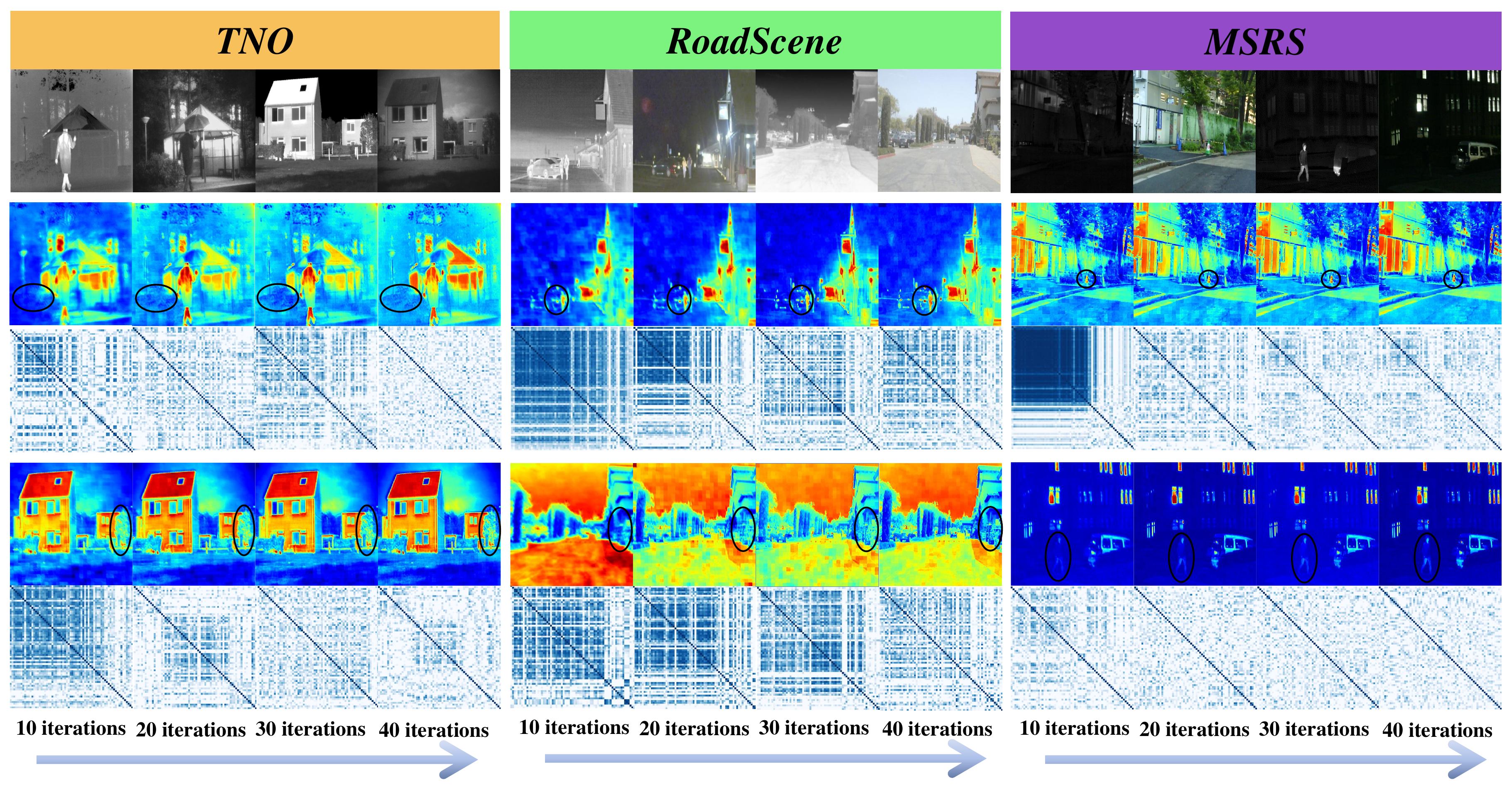}
    \caption{Visualization of training results at different iterations. The first row represents the source image. The second and fourth rows show intermediate fusion results, while the third and fifth rows display the evolution of Riemannian attention weight matrices. Initially, the weights concentrate on intra-modality (diagonal regions), yielding blurry fused outputs. As training progresses, the weights gradually spread to inter-modality (off-diagonal regions), demonstrating the network’s ability to integrate cross-modal information. This evolution produces more structurally coherent fused images.}
    \label{Visualization}
\end{figure*}

In order to allow the model to extract and learn higher-level feature representations from the input data, consecutive convolutional layers and LeakyReLU activation functions are added after obtaining the feature map by SPDAM. At the same time, by using different filters to extract features at various levels, higher-level abstract representations are established, and deep features of the two modalities are fused and decoded:
\begin{equation}
    I_{\text{f}}=conv_{3\times3}\left ( cat\left ( F_{\text{ir}}, F_{\text{vi}} \right ) \right ),
    \label{Conv}
\end{equation}
where \(I_{\text{f}}\) denotes the final fusion result obtained through the above steps.

\subsection{Cross Modal SPD Fusion Strategy}\label{cross}
The core of this strategy is to use a structured feature map stacking method, in which the feature maps of two different modalities are arranged vertically for interaction within the same dimensional space. Specifically, the two source images are firstly divided into patches and flattened into vectors as shown in Fig. \ref{Cross fusion strategy}. In our work, the patch size, the stride between blocks, and the overlapping block size are set to 16, 8, and 8, respectively. After the patches are flattened, we obtain the row vector matrices \(\mathbf{M}_{\text{ir}}\) and \(\mathbf{M}_{\text{vi}}\), corresponding to different modalities. The process of vertical arrangement can be described as constructing a new stacked matrix \(\mathbf{M}_{\text{stacked}}\), which merges \(\mathbf{M}_{\text{ir}}\) and \(\mathbf{M}_{\text{vi}}\) as follows:
\begin{equation}
    \mathbf{M}_{\text{stacked}}=\begin{bmatrix}
 \mathbf{M}_{\text{ir}}\\
\mathbf{M}_{\text{vi}}
\end{bmatrix},
\label{Stack}
\end{equation}
where \(\mathbf{M}_{\text{ir}} \in \mathbb{R}^{m\times n}\), \(\mathbf{M}_{\text{vi}} \in \mathbb{R}^{p\times n}\) and \(\mathbf{M}_{\text{stacked}} \in \mathbb{R}^{\left ( m+p \right ) \times n}\). Here \(m\) and \(p\) denote the number of the row vectors for the two modalities, and \(n\) represent the dimension of the row vector. For any two rows \(\mathbf{M}_{i}\) and \(\mathbf{M}_{j}\) \(\left ( i,j=1,2,\dots,m+p,i\ne j\right)\) in \(\mathbf{M}_{\text{stacked}}\), the covariance representation between the row vectors can be formulated as:
{
\small
\begin{equation}
    Cov\left (\mathbf{M}_{i},\mathbf{M}_{j}\right )=\frac{1}{n-1} \sum_{k=1}^{n} \left (\mathbf{M}_{ik}-\bar{\mathbf{M}}_{i}\right ) \left (\mathbf{M}_{jk}-\bar{\mathbf{M}}_{j}\right )^{T},
    \label{Cov}
\end{equation}
}
where \(\bar{\mathbf{M}}_{i}\) and \(\bar{\mathbf{M}}_{j}\) represent the average of all observed feature values for the \(i\)-th patch and the \(j\)-th patch, respectively.

Then, a composite covariance matrix is constructed to analyze the collaborative variation trends between composite data points on the high-dimensional data manifold. The matrix is divided into four quadrant blocks:
\begin{equation}
    \mathbf{C}_{\mathbf{M}} = \begin{bmatrix}
        \mathbf{C}_{\mathbf{M}_{\text{ir}}} & \mathbf{C}_{\mathbf{M}_{\text{ir}}\mathbf{M}_{\text{vi}}} \\ 
        \mathbf{C}_{\mathbf{M}_{\text{vi}}\mathbf{M}_{\text{ir}}} & \mathbf{C}_{\mathbf{M}_{\text{vi}}}
    \end{bmatrix},
    \label{CX}
\end{equation}
where
\begin{equation}
    \mathbf{C}_{\mathbf{M}_{\text{ir}}} = \frac{1}{n-1}\sum_{k=1}^{n} \left( \mathbf{M}_{\text{ir}_{k}} - \bar{\mathbf{M}}_{\text{ir}} \right) \left( \mathbf{M}_{\text{ir}_{k}} - \bar{\mathbf{M}}_{\text{ir}} \right)^{T},
    \label{CXir}
\end{equation}
and
\begin{equation}
    \mathbf{C}_{\mathbf{M}_{\text{vi}}} = \frac{1}{n-1}\sum_{k=1}^{n} \left( \mathbf{M}_{\text{vi}_{k}} - \bar{\mathbf{M}}_{\text{vi}} \right) \left( \mathbf{M}_{\text{vi}_{k}} - \bar{\mathbf{M}}_{\text{vi}} \right)^{T}.
    \label{CXvi}
\end{equation}
Here, \(\mathbf{C}_{\mathbf{M}_{\text{ir}}}\) and \(\mathbf{C}_{\mathbf{M}_{\text{vi}}}\) are the covariance matrices of the patches within \(\mathbf{M}_{\text{ir}}\) and \(\mathbf{M}_{\text{vi}}\), respectively, representing the self-correlation characteristics of each modality. The cross-modal covariance is given by
{
\small
\begin{equation}
\label{CXirvi}
\begin{split}
    \mathbf{C}_{\mathbf{M}_{\text{ir}}\mathbf{M}_{\text{vi}}}&=\mathbf{C}_{\mathbf{M}_{\text{vi}}\mathbf{M}_{\text{ir}}}\\&=\frac{1}{n-1}\sum_{k=1}^{n}\left ( \mathbf{M}_{\text{ir}_{k}}-\bar{\mathbf{M}}_{\text{ir}} \right )\left ( \mathbf{M}_{\text{vi}_{k}}-\bar{\mathbf{M}}_{\text{vi}} \right )^{T}, 
\end{split}
\end{equation}
}
where \(\mathbf{C}_{\mathbf{M}_{\text{ir}}\mathbf{M}_{\text{vi}}}\) and \(\mathbf{C}_{\mathbf{M}_{\text{vi}}\mathbf{M}_{\text{ir}}}\) capture the relationship between the two modalities, with \(\mathbf{C}_{\mathbf{M}_{\text{vi}}\mathbf{M}_{\text{ir}}}\) being the transpose of \(\mathbf{C}_{\mathbf{M}_{\text{ir}}\mathbf{M}_{\text{vi}}}\).

This Riemannian representation fundamentally differs from Euclidean feature addition, as trivial summation fails to capture higher-order statistical interactions. In contrast, \(\mathbf{C}_{\mathbf{M}}\) preserves the complex nonlinear dependencies between modalities through its tensor structure. During backpropagation, the gradients constrained by the manifold geometry adaptively refine \(\mathbf{C}_{\mathbf{M}}\), optimizing the intrinsic connections between modalities. This process yields enhanced features \(\mathrm{IR}\) and \(\mathrm{VI}\) with geometrically consistent fusion results.

\subsection{Manifold Learning Network}
To facilitate the understanding, we first provide a brief introduction to SPD Riemannian geometry. Firstly, the SPD network is strictly defined on the SPD manifold \(Sym_{++}^{d_{k}}\), which is formed by a set of \(d_{k}\)-dimensional SPD matrices. These matrices are used to capture statistical correlations between different modes. Here, \(X_{k-1} \in \mathbb{R}^{d_{k-1}\times d_{k-1}} \) denotes the input SPD matrix of the \(k\)-th layer.

As shown in Fig. \ref{Framework}(b),the process of SPD matrix learning starts with a bidirectional mapping layer, known as the BiMap layer, which encodes the global structure of \(X_{0}\) as well as the complex inter-modal correlations. Through the bidirectional linear mapping, the high-dimensional SPD matrix is gradually transformed into a more compact, lower-dimensional representation, denoted as \(X_{1}\), aiming to merge the fine-grained interactions between different modes into a more discriminative low-dimensional SPD manifold.

Subsequently, the ReEig layer obtains a new representation \(X_{2}\) by performing a nonlinear transformation based on eigenvalue regularization. The key function of this layer is to endow the network with the capability of nonlinear learning while preserving the Riemannian manifold properties of the data. This significantly contributes to the facilitation of more effective SPD parameter learning to a certain extent.

After a series of BiMap and ReEig operations, a set of SPD matrices are obtained which inherently maintain their Riemannian geometric characteristics, and through the logarithmic operation of the LogEig layer, data points on the SPD manifold are mapped to the tangent space at the identity matrix \(T_{I}Sym_{++}^{d_{k}}\) to facilitate traditional Euclidean computations. Then, not only are the cross-modal features integrated on the manifold, but the phase information of the SPD matrices is also preserved. Fig. \ref{Visualization} displays the SPD matrix during the network training.

Accompanied this learning process, the neural network continuously performs gradient descent on the complex Riemannian manifold. Utilizing the training-derived SPD manifold weight parameters, we guide the extraction of highly correlated non-Euclidean features, thereby facilitating the cross-fusion of different modalities.

\begin{figure}
    
    \centering
    % \captionsetup{justification=justified,singlelinecheck=false}
    \includegraphics[width=1\linewidth]{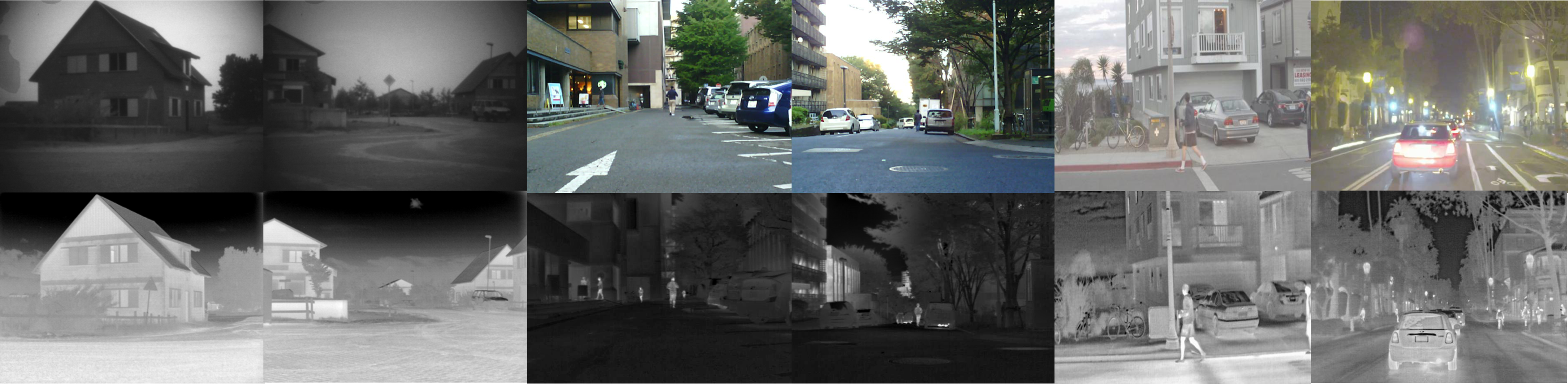}

    \begin{minipage}{0.15\textwidth}
        \centering
        TNO
    \end{minipage}
    \hspace{1pt} % 文本之间的空格
    \begin{minipage}{0.15\textwidth}
        \centering
        MSRS
    \end{minipage}
    \hspace{1pt} % 文本之间的空格
    \begin{minipage}{0.15\textwidth}
        \centering
        RoadScene
    \end{minipage}
    \hspace{1pt} % 文本之间的空格

    \caption{The examples of three datasets: TNO, MSRS and RoadScene.}
    \label{Dataset}
\end{figure}

\begin{figure*}[ht!]
    \small
    \centering
    % \captionsetup{justification=justified,singlelinecheck=false}
    % \begin{subfigure}[b]{1\linewidth}
        \includegraphics[width=1\linewidth]{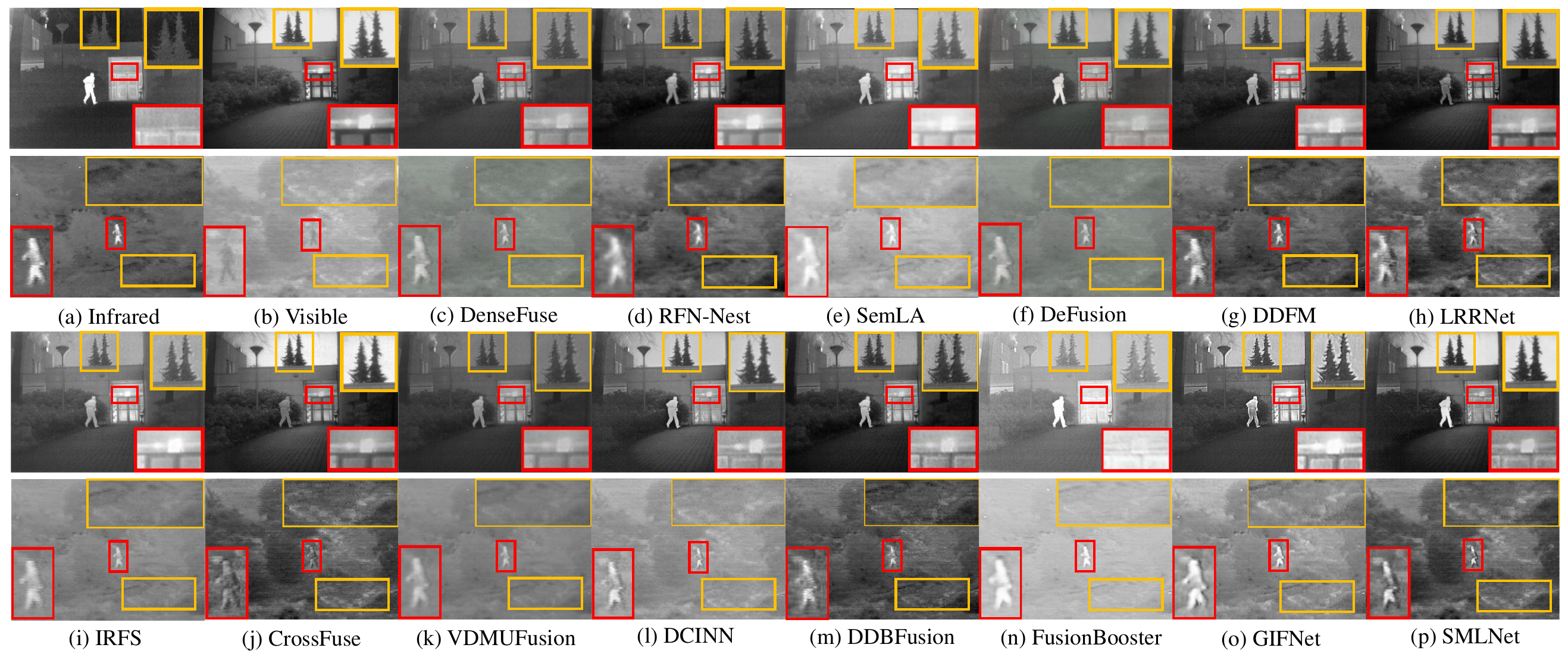}
    
    \caption{ Infrared and visible image fusion experiment on TNO dataset.}
    \label{TNO}
\end{figure*}

\begin{figure*}[ht!]
    \small
    \centering
    % \captionsetup{justification=justified,singlelinecheck=false}
    % \begin{subfigure}[b]{1\linewidth}
        \includegraphics[width=1\linewidth]{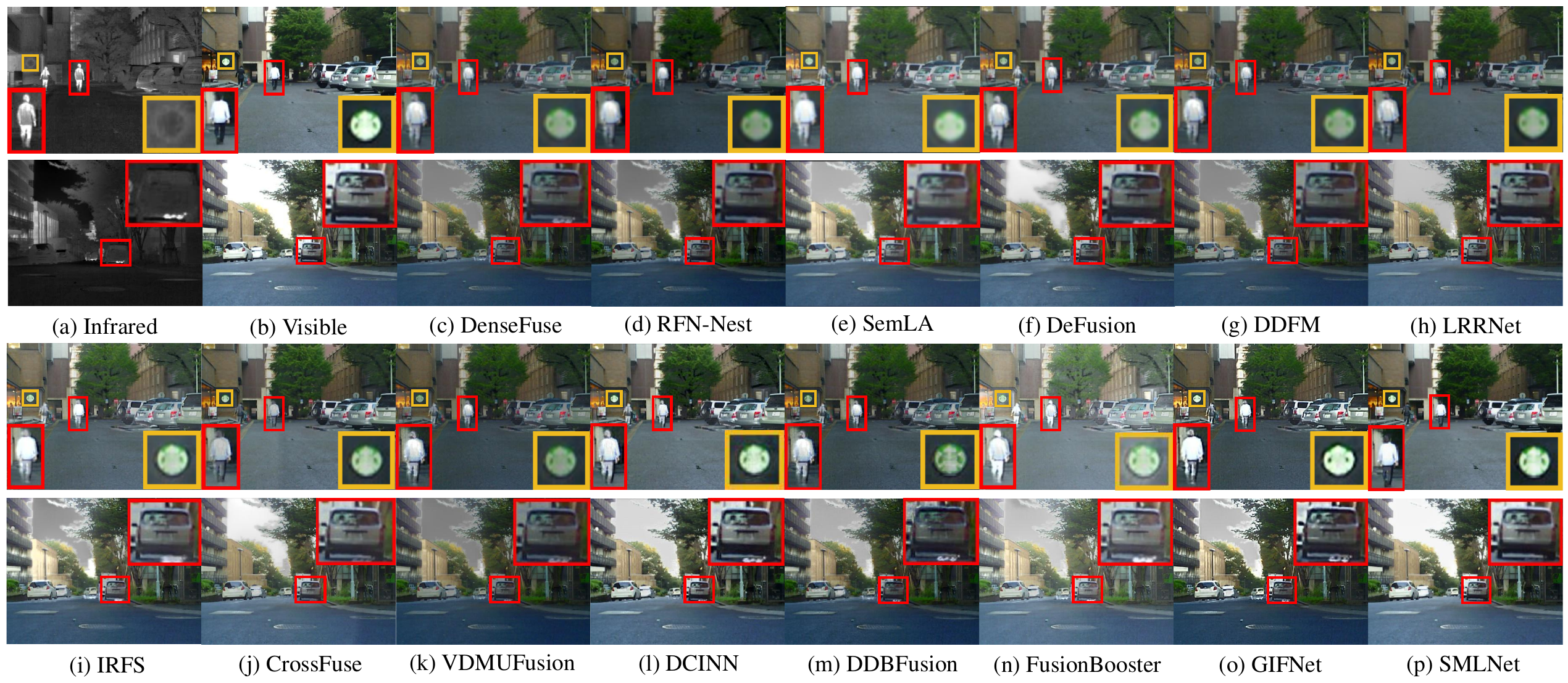}
    % \end{subfigure}

    \caption{ Infrared and visible image fusion experiment on MSRS dataset.}
    \label{MSRS}
\end{figure*}

\begin{figure*}[ht!]
    \small
    \centering
    % \captionsetup{justification=justified,singlelinecheck=false}
    % \begin{subfigure}[b]{1\linewidth}
        \includegraphics[width=1\linewidth]{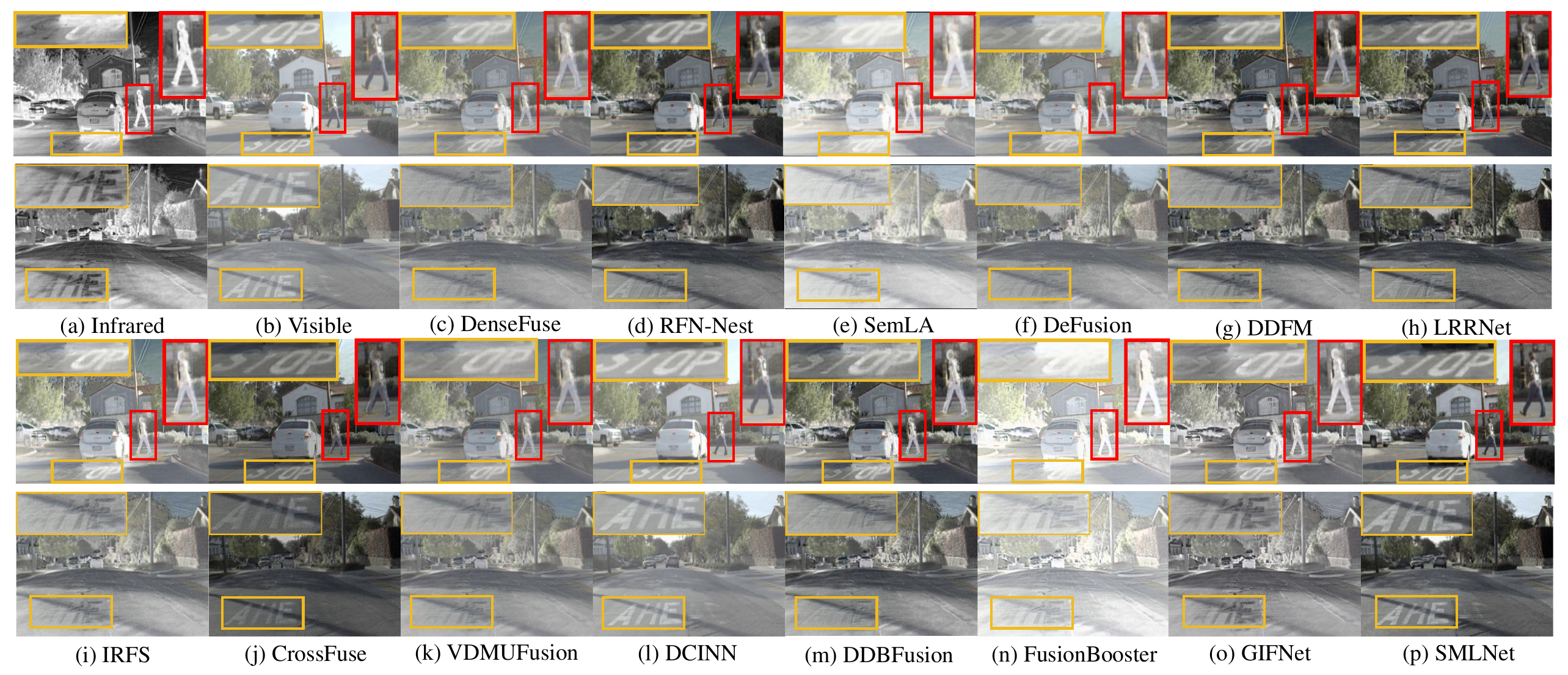}
    
    \caption{ Infrared and visible image fusion experiment on RoadScene dataset.}
    \label{RoadScene}
\end{figure*}

\subsection{SPD Manifold Attention Module}
Following conventional spatial attention, we introduce the Symmetric Positive Definite Attention Module (SPDAM) to exploit global semantic associations within Riemannian manifolds. Central to SPDAM is its covariance matrix, which inherently encapsulates symmetry and positive definiteness while enabling manifold attention redistribution in Euclidean space. This mechanism effectively ``flattens" points on the SPD manifold through a mapping function \(f: \mathcal{M} \rightarrow \mathbb{R}^d\), linearly approximating manifold structures within Euclidean coordinates. 

Crucially, as shown in Fig. \ref{Framework}(c), each element in our SPD feature matrix quantifies statistical correlations between the \(i\)-th and \(j\)-th patches. Based on this, we integrate cross-modal statistical correlation weights by multiplying the learned weight matrix with blocked row vectors. This operation specifically weights the \(j\)-th pixel position (corresponding locations across all patches) into the \(i\)-th patch's features, thereby restructuring local features within a global context. Furthermore, we learn weight parameters directly in the manifold space, yielding more principled weighting schemes. Later ablation studies empirically validate this design advantage.

% Specifically, the interaction, as depicted in the covariance matrix entries post-SPDAM application, underscores the statistical correlation between distinct patches within an image, enriched by the SPD weight matrix's interaction with the blocked row vector matrix. This interaction not only aggregates the variance weights between image patches but also achieves an effective approximation within the Euclidean space by linearizing the manifold.

Through this natural weighting transition, the spatial features on the manifolds of infrared and visible light modalities are extracted and enhanced, which further improves the fusion network's capability in representing global and local associative information.

\subsection{Loss Function}
Our SMLNet is constrained by int loss \(L_{\text{int}}\), grad loss \(L_{\text{grad}}\) , structure similarity loss \(L_{\text{ssim}}\) and covariance loss \(L_{\text{cov}}\). This is done to enable the network to further enhance semantic information through the correlation between deep features, while learning the color brightness information and texture details in the image. The total loss function \(L_{\text{total}}\) is defined as follows:
\begin{equation}
    L_{\text{total}}=L_{\text{int}}+\alpha L_{\text{grad}}+\beta L_{\text{ssim}}+\gamma L_{\text{cov}},
    \label{Ltotal}
\end{equation}
where \(L_{\text{int}}\) calculates the difference between the generated image and the target image at the pixel level. It reflects the difference in intensity between images, and is defined as:
\begin{equation}
    L_{\text{int}}=\frac{1}{HW}\parallel I_{\text{f}}-max(I_{\text{ir}},I_{\text{vi}}) \parallel _{1},
    \label{Lint}
\end{equation}
where \(H\) and \(W\) respectively represent the height and width of the image, \(max\left ( \cdot\right )\) signifies taking the maximum value of each element, and \(\parallel \cdot \parallel _{1}\) is \(l_{1}-norm\).

At the same time, we expect the fusion result to contain more detailed textures, and constrain the image from the changes in gradient, as shown below:
\begin{equation}
    L_{\text{grad}}=\frac{1}{HW}\parallel \left | \nabla I_{\text{f}} \right | -max(\left | \nabla I_{\text{ir}} \right | , \left | \nabla I_{\text{vi}} \right | ) \parallel _{1},
    \label{Lgrad}
\end{equation}
where \(\nabla\) is the Sobel operator, and \(|\cdot|\) represents the absolute value operation.

The SSIM loss is a loss function based on the Structural Similarity Index Measure (SSIM), which is used to measure the similarity of two images. Unlike pixel loss, which directly compares pixel values, SSIM considers three relatively independent dimensions: brightness, contrast, and structure. This is closer to the perceptual characteristics of the human visual system. It is defined as:
\begin{equation}
    L_{\text{ssim}}=\left (1-ssim \left( I_{\text{f}},I_{\text{vi}} \right ) \right )+\left (1-ssim\left (I_{\text{f}},I_{\text{ir}}\right )\right ),
    \label{Lssim}
\end{equation}
where \(ssim \left (\cdot \right )\) represents the structural similarity between two images. The larger it is, the higher the structural similarity between the fused image and the source image.

Furthermore, to restrain the highly correlated features of the image, we employ VGG-16 ~\citep{simonyan2014very}, trained on ImageNet, to extract features. The third and fourth layers of the VGG network are selected to devise the loss function, which allows us to capture a more abstract characterization of the correlations. \(L_{\text{cov}}\) is defined as follows:
\begin{equation}
    L_{\text{cov}}=\sum_{k=3}^{4} || Cov(\phi (I_{\text{f}})^{k})-Cov(\phi (I_{\text{ir}})^{k})||_{1},
    \label{Lcov}
\end{equation}
where \(Cov \left (\cdot \right )\) denotes the covariance matrix of the feature map. Its diagonal elements represent the auto-correlation of each feature appearing in the image, while the off-diagonal elements indicate the correlations between different features. The \(k\) denotes the network depth.

\begin{table*}[ht!]
\small
\centering % Centering the table

% \captionsetup{justification=justified,singlelinecheck=false}
\caption{Quantitative Experiments on the TNO Dataset. We highlight the top three metrics using \textbf{\textsl{bolditalic}}, \textbf{bold} and \textsl{italic} fonts.}
\label{Quantitave TNO}
% Table caption
\begin{tabular}{ccccccccc} % There are 8 centered columns (c), you can change to l (left align) or r (right align) as needed
\toprule
Methods & Year & EN & MI & SD & SF & VIF & AG & \(Q^{AB/F}\)  \\
\midrule
 DenseFuse~\citep{li2018densefuse} & 2018 & 6.380 & 2.227  & 25.259  & 6.570 & 0.574 & 2.569 & 0.349   \\
 RFN-Nest~\citep{11} & 2021 & 6.971 & 2.126 & 37.090 & 5.931 & 0.550 & 2.691 & 0.333\\
 DeFusion~\citep{liang2022fusion} & 2022 & 6.607 & \textsl{2.627} & 30.868 & 6.461 & 0.559 & 2.624 & 0.365 \\
  SemLA~\citep{xie2023semantics} & 2023 & 6.879 & 2.051 & 37.890 & 8.528 & 0.470 & 2.892 & 0.265 \\
   DDFM~\citep{zhao2023ddfm} & 2023 & 6.854 & 2.231 & 34.418 & 8.629 & \textsl{0.631} & 3.405 & \textbf{0.434} \\
 LRRNet~\citep{8} & 2023 & \textbf{6.991} & 2.518 & \textsl{40.984} & 9.608 & 0.548 & 3.790 & 0.352 \\
 IRFS~\citep{wang2023interactively} & 2023 & 6.651 & 2.150 & 31.784 & 8.939 & 0.587 & 3.242 & 0.397 \\
 CrossFuse~\citep{li2024crossfuse} & 2024 & 6.908 & \textbf{2.952} & 39.976 & 9.951 & \textbf{0.716} & 3.733 & \textsl{0.425} \\
 VDMUFusion~\citep{shi2024vdmufusion} & 2024 & 6.358 & 2.076 & 24.980 & 4.832 & 0.502 & 2.048 & 0.249 \\
 DCINN~\citep{2024dcinn} & 2024 & 6.772 & 2.143 & 34.977 & 9.626 & 0.364 & 3.837 & 0.290 \\
 DDBFusion~\citep{zhang2025ddbfusion} & 2025 & 6.620 & 1.935 & 37.343 & 10.585 & 0.511 & \textsl{4.143} & 0.350 \\
 FusionBooster~\citep{cheng2024fusionbooster} & 2025 & 6.516 & 2.455 & 32.863 & \textsl{10.898} & 0.480 & 3.455 & 0.319 \\
 GIFNet~\citep{cheng2025cvpr_gifnet} & 2025 & \textsl{6.974} & 1.990 & \textbf{41.429} & \textbf{\textsl{13.742}} & 0.501 & \textbf{\textsl{5.132}} & 0.347 \\
 \midrule
 SMLNet & Ours & \textbf{\textsl{7.068}} & \textbf{\textsl{3.894}} & \textbf{\textsl{43.209}} & \textbf{11.098} & \textbf{\textsl{0.832}} & \textbf{4.301}  & \textbf{\textsl{0.516}}  \\
\bottomrule
\end{tabular}
\end{table*}

\begin{table*}[ht!]
\small
\centering % Centering the table

% \captionsetup{justification=justified,singlelinecheck=false}

\caption{Quantitave Experiments on the MSRS Dataset. The top three metrics are displayed in \textbf{\textsl{bolditalic}}, \textbf{bold} and \textsl{italic} fonts, respectively.}
\label{Quantitave MSRS}
% Table caption
\begin{tabular}{ccccccccc} % There are 8 centered columns (c), you can change to l (left align) or r (right align) as needed
\toprule
Methods & Years & EN & MI & SD & SF & VIF & AG & \(Q^{AB/F}\)  \\
\midrule
DenseFuse~\citep{li2018densefuse} & 2018 & 5.931 & 2.666  & 23.550  & 6.020 & 0.692 & 2.053 & 0.368  \\

RFN-Nest~\citep{11} & 2021 & 6.196 & 2.460 & 29.078 & 6.163 & 0.656 & 2.115 & 0.390 \\
DeFusion~\citep{liang2022fusion} & 2022 & 6.383 & 2.990 & 35.429 & 8.146 & 0.730 & 2.654 & 0.507 \\
SemLA~\citep{xie2023semantics} & 2023 & 6.423 & 2.461 & 33.122 & 6.351 & 0.617 & 2.257 & 0.290 \\
DDFM~\citep{zhao2023ddfm} & 2023 & 6.175 & 2.735 & 28.925 & 7.388 & \textsl{0.743} & 2.522 & 0.474 \\
LRRNet~\citep{8} & 2023 & 6.192 & 2.928 & 31.758 & 8.473 & 0.541 & 2.651 & 0.454 \\
IRFS~\citep{wang2023interactively} & 2023 & \textbf{\textsl{6.603}} & 2.159 & 35.868 & 9.888 & 0.735 & \textsl{3.155} & 0.477 \\
CrossFuse~\citep{li2024crossfuse} & 2024 & \textbf{6.494} & \textbf{3.132} & \textsl{36.447} & 9.661 & \textbf{0.839} & 3.019 & \textbf{0.561} \\
VDMUFusion~\citep{shi2024vdmufusion} & 2024 & 5.948 & 2.331 & 23.600 & 6.490 & 0.633 & 2.242 & 0.344 \\
DCINN~\citep{2024dcinn} & 2024 & 6.044 & \textsl{3.054} & \textbf{40.288} & \textbf{10.783} & 0.721 & \textbf{\textsl{3.473}} & \textsl{0.509} \\
DDBFusion~\citep{zhang2025ddbfusion} & 2025 & 5.971 & 2.210 & 28.422 & 8.551 & 0.628 & 2.791 & 0.385 \\
FusionBooster~\citep{cheng2024fusionbooster} & 2025 & 6.316 & 2.075 & 30.215 & 8.801 & 0.639 & 3.051 & 0.422 \\
GIFNet~\citep{cheng2025cvpr_gifnet} & 2025 & 5.940 & 1.971 & 32.900 & \textbf{\textsl{12.705}} & 0.582 & \textbf{3.367} & 0.416 \\
\midrule
SMLNet & Ours & \textsl{6.478} & \textbf{\textsl{3.903}} & \textbf{\textsl{41.571}} & \textsl{9.965} & \textbf{\textsl{0.855}} & 3.063 & \textbf{\textsl{0.566}} \\
\bottomrule
\end{tabular}
\end{table*}

\begin{table*}[ht!]
\small
\centering % Centering the table

% \captionsetup{justification=justified,singlelinecheck=false}

\caption{Quantitave Experiments on the RoadScene Dataset. For the best three metrics, we use \textbf{\textsl{bolditalic}}, \textbf{bold} and \textsl{italic} fonts to distinguish them.}
\label{Quantitave RoadScene}
% Table caption
\begin{tabular}{ccccccccc} % There are 8 centered columns (c), you can change to l (left align) or r (right align) as needed
\toprule
Methods & Years & EN & MI & SD & SF & VIF & AG & \(Q^{AB/F}\)  \\
\midrule
DenseFuse~\citep{li2018densefuse} & 2018 & 6.765 & 2.939 & 31.384 & 5.211 & 0.610 & 2.226 & 0.389  \\
RFN-Nest~\citep{11} & 2021 & \textbf{7.253} & 2.877 & 44.926 & 6.641 & \textbf{0.717} & 2.733 & 0.375 \\
DeFusion~\citep{liang2022fusion} & 2022 & 6.906 & 3.092 & 34.790 & 5.437 & 0.601 & 2.323 & 0.412 \\
SemLA~\citep{xie2023semantics} & 2023 & 6.954 & 2.856 & 38.566 & 8.119 & 0.564 & 2.515 & 0.271 \\
DDFM~\citep{zhao2023ddfm} & 2023 & \textsl{7.242} & 2.952 & 43.262 & 7.223 & 0.676 & 3.080 & \textbf{\textsl{0.483}} \\
LRRNet~\citep{8} & 2023 & 7.145 & 2.866 & 44.110 & 8.196 & 0.578 & \textsl{3.299} & 0.382 \\
IRFS~\citep{wang2023interactively} & 2023 & 7.010 & 2.999 & 37.631 & 6.385 & 0.649 & 2.695 & \textsl{0.476} \\
CrossFuse~\citep{li2024crossfuse} & 2024 & 7.189 & \textbf{3.520} & \textbf{47.273} & 7.578 & 0.663 & 3.055 & 0.366 \\
VDMUFusion~\citep{shi2024vdmufusion} & 2024 & 6.798 & 2.845 & 32.242 & 5.665 & 0.576 & 2.358 & 0.375 \\
DCINN~\citep{2024dcinn} & 2024 & 6.940 & \textsl{3.313} & 36.681 & 7.526 & \textsl{0.706} & 3.046 & \textbf{0.480} \\
DDBFusion~\citep{zhang2025ddbfusion} & 2025 & 6.880 & 2.723 & 44.499 & \textbf{8.668} & 0.633 & \textbf{3.597} & 0.438 \\

FusionBooster~\citep{cheng2024fusionbooster} & 2025 & 6.524 & 3.127 & 29.351 & 6.098 & 0.495 & 2.519 & 0.347 \\
GIFNet~\citep{cheng2025cvpr_gifnet} & 2025 & 7.241 & 2.659 & \textsl{46.338} & \textbf{\textsl{11.460}} & 0.576 & \textbf{\textsl{4.293}} & 0.413 \\

\midrule
SMLNet & Ours & \textbf{\textsl{7.270}} & \textbf{\textsl{5.055}} & \textbf{\textsl{56.767}} & \textsl{8.549} & \textbf{\textsl{1.104}} & 3.139 & 0.415 \\
\bottomrule
\end{tabular}
\end{table*}

\section{Experiments and Analyses}
This section presents fusion results and systematically evaluates SMLNet. We first detail the experimental setup, then conduct ablation studies to validate the effectiveness of both the SPD attention module and fusion strategy. Comprehensive benchmarks on image fusion tasks and downstream applications, supported by qualitative and quantitative analyses, demonstrate the superiority of our method. Efficiency comparisons further highlight the advantages of our lightweight design. To ensure a thorough assessment, we also perform analysis of the method's performance under extreme scenarios.
\subsection{Experimental Settings}
In this section, we discuss the datasets, parameter settings, comparative methods, and quantitative metrics used in our approach respectively.
\subsubsection{Datasets}
In our work, we selected 1083 pairs of corresponding infrared and visible images from the MSRS dataset ~\citep{tang2022piafusion} as training data, ensuring a diverse representation of multi-modal scenarios. The size of the training images is standardized to 256×256 pixels. During the testing phase, we use 40 pairs of images from TNO ~\citep{toet2012progress}, 361 pairs of images from MSRS and 40 pairs of images from RoadScene~\citep{xu2020u2fusion} as the test sets, respectively. The dimensions of the test images are typically not fixed, and some instances from the dataset are shown in Fig. \ref{Dataset}.

\subsubsection{Parameter Setting}
We implemented the algorithm using Pytorch. In the training phase, an end-to-end strategy was employed to train the model on an NVIDIA TITAN RTX GPU. Different optimizers were set for the manifold module and the convolutional layers. Within the manifold module, the SGD optimizer on the Stiefel manifold was used to update the weights of the BiMap layer, with a learning rate set to 0.01. For the convolutional layers, we used the Adam optimizer to update the weights, with the learning rate for this part set to 0.0001.

At the same time, in our loss function, the parameters \(\alpha\), \(\beta\), \(\gamma\) are set to 1, 10, 20, respectively, to achieve the best fusion effect.

\subsubsection{The Methods Compared and the Quality Metrics Used}
The method presented in this article was compared and evaluated with thirteen different state-of-the-art image fusion network approaches, including some classic and latest methods. These are: DenseFuse~\citep{li2018densefuse}, RFN-Nest~\citep{11}, DeFusion~\citep{liang2022fusion}, SemLA~\citep{xie2023semantics}, DDFM~\citep{zhao2023ddfm} , LRRNet~\citep{8}, IRFS~\citep{wang2023interactively}, CrossFuse~\citep{li2024crossfuse}, VDMUFusion~\citep{shi2024vdmufusion}, DCINN~\citep{2024dcinn}, DDBFusion~\citep{zhang2025ddbfusion}, FusionBooster~\citep{cheng2024fusionbooster} and GIFNet~\citep{cheng2025cvpr_gifnet}. Regarding the quality metrics, seven indices were chosen for performance evaluation, including: Entropy (En), Mutual Information (MI), Standard Deviation (SD), Spatial Frequency (SF), Visual Information Fidelity (VIF), Average Gradient (AG), \(Q^{AB/F}\).These indices measure the overall quality of images from different perspectives, and the descriptions of the metrics can be found in~\citep{ma2019infrared}.

% \begin{table}[h]
% \centering
% \caption{Your Table Caption}
% \label{your-label}
% \begin{tabular}{@{}llr@{}}
% \toprule
% Block_num & En & SD & SF & SCD\\ \midrule
% 0   & 6.705   & 27.984   & 5.093   & 1.820\\
% 2   & 6.925   & 35.698   & 5.988   & 1.804\\
% 3   & 6.993   & 39.513   & 6.298   & 1.765 \\
% 4   & 7.015   & 41.226   & 6.373   & 1.719\\
% 5   & 7.037   & 42.461   & 6.437   & 1.678\\ \bottomrule
% \end{tabular}
% \end{table}

\subsection{Comparative  Results Analysis}
In this section, we carry out qualitative and quantitative experiments with the proposed SMLNet on the aforementioned three classical infrared-visible datasets to verify the method performance.

\begin{figure}
    \centering
    % \captionsetup{justification=justified,singlelinecheck=false}
    \includegraphics[width=1\linewidth]{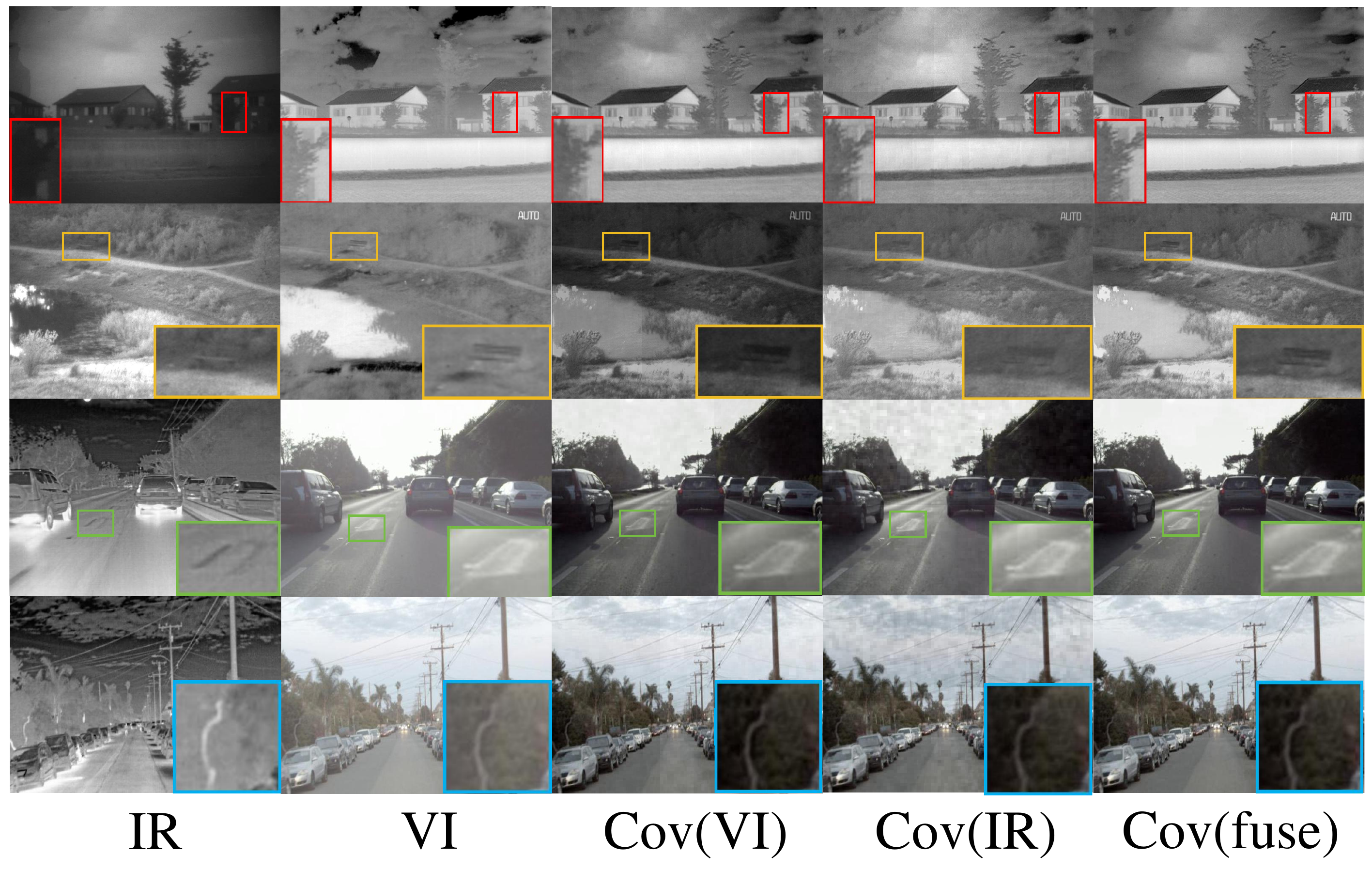}

    \caption{The qualitative results obtained using single-modal and multi-modal fusion strategies. When the second-order statistical information of the infrared modality is disregarded, the fusion results appear darker with lower contrast. Conversely, excluding the visible modality from covariance computation leads to blurred outcomes.}
    \label{Ablation1}
\end{figure}

\subsubsection{The Fusion Results on TNO}
First, we conduct experiments on the TNO dataset, which are shown in Fig. \ref{TNO}.

It is worth mentioning that for the parts of the image involving the tree tops and the sky, as indicated by the yellow highlighted areas, methods such as (c), (d), (f) and (k) are unable to maintain the original color features and texture details of the images. In comparison, although (e), (i), (m), (n) and (o) have improvements in terms of visible light brightness and detail, their ability to preserve fundamental objection contours remains an issue, and they also introduce relatively unfriendly overexposure. Meanwhile, (g), (h), (j) and (l), perform well in these aspects, but it is still difficult to achieve a balance between capturing high correlation scene information and obtaining low correlation semantic information in different modalities. Our SMLNet enhanced the clarity and brightness of the image in areas where the background is more indistinct. On the other hand, in the parts concerning the targeted individuals, as shown by the red highlighted areas, our approach was able to offer clearer delineation of the edge regions, resulting in higher contrast, while also conforming to human visual perception.

Additionally, we utilized the previous seven indices to conduct an objective evaluation of the methods mentioned above, as shown in the Table \ref{Quantitave TNO}. Among the various state-of-the-art fusion methods, SMLNet achieved the highest level in five classic image fusion metrics (EN, MI, SD, VIF, \(Q^{AB/F}\)) and reached the second-highest level in the remaining two metric (SF and AG). These results indicate that our method not only enhances the overall sharpness of the image but also presents a more balanced and realistic visual effect, significantly surpassing existing technologies in terms of visual quality.

\begin{figure}
    \small
    \centering
    % \captionsetup{justification=justified,singlelinecheck=false}
    \includegraphics[width=1\linewidth]{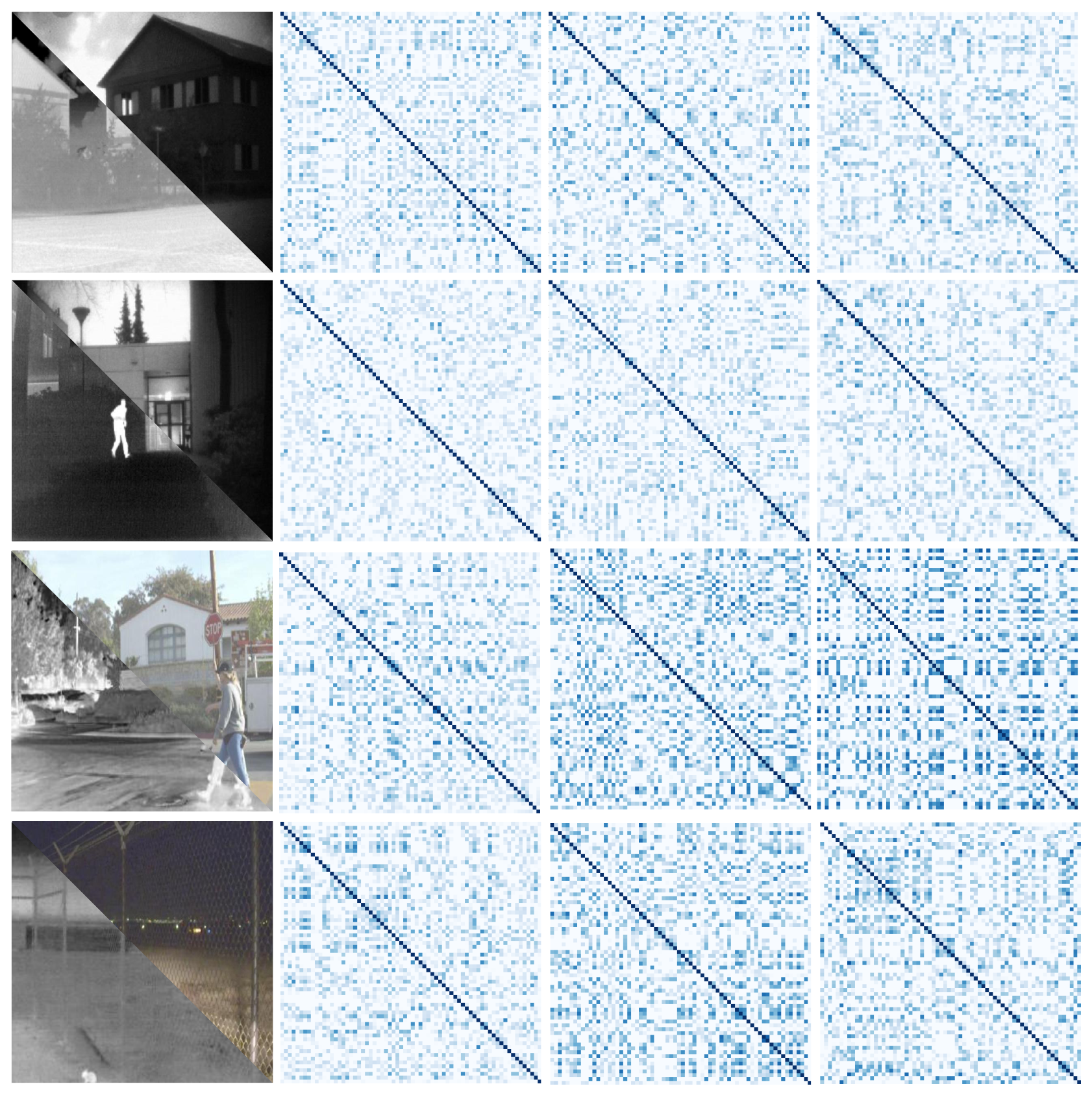}

    \begin{minipage}{0.11\textwidth}
        \centering
        IR/VI
    \end{minipage}
    \hspace{1pt} % 文本之间的空格
    \begin{minipage}{0.11\textwidth}
        \centering
        b1-r1
    \end{minipage}
    \hspace{1pt} % 文本之间的空格
    \begin{minipage}{0.11\textwidth}
        \centering
        b2-r2
    
    \end{minipage}
    \hspace{1pt} % 文本之间的空格
    \begin{minipage}{0.11\textwidth}
        \centering
        b3-r3
        
    \end{minipage}

    \caption{Visualization results of intermediate covariance matrices obtained from different manifold layer structures. The size of each covariance matrix is 1089×1089.}
    \label{Ablation2}
\end{figure}

\subsubsection{Fusion Results on MSRS}
To verify the generalization performance of our SMLNet, we selected more IR-VI pairs from the MSRS dataset for analysis and comparison, as shown in Fig. \ref{MSRS}. Compared to methods like (d), (e), (f), (i), (l), (n) and (o), our SMLNet method highlights infrared targets and performs visual enhancement more distinctively, as illustrated by the red boxed sections in the images. This, to some extent, reflects that the Riemannian attention mechanism can effectively capture deep semantic information in the infrared modality. At the same time, fused images generated by (c), (g), (h), (j), (k) and (m) appear relatively blurry, whereas our method retains more detailed parts of the objects within the scene (as indicated by the yellow boxed sections in the images). This means that our method preserves not only the prominent features from the infrared image but also the texture details from the visible light image, which is crucial for the task of image fusion.

In the Table \ref{Quantitave MSRS}, compared to other fusion methods, our proposed SMLNet achieved the best values in four metrics (MI, SD, VIF and \(Q^{AB/F}\)) competitive results in two others (EN and SF). This indicates that the images fused via SMLNet contain more edge details and have a higher degree of information fidelity, also reflecting our work's success in achieving cross-modal pixel and semantic-level fusion.

\begin{figure}
    \centering
    % \captionsetup{justification=justified,singlelinecheck=false}
    % \begin{subfigure}[b]{1\linewidth}  
        \includegraphics[width=\linewidth]{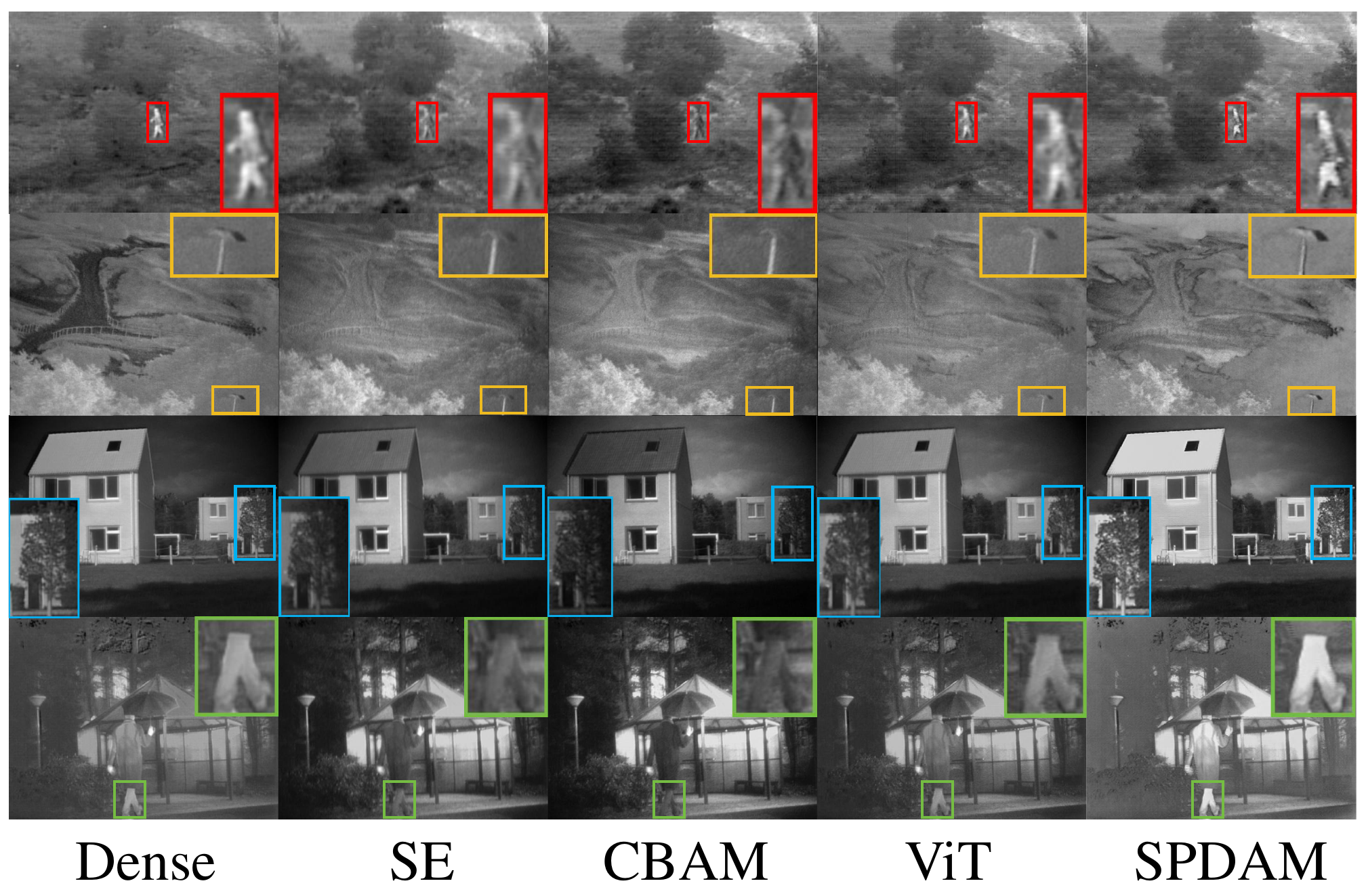}
 
    % \end{subfigure}
    
    \caption{ Conventional attention networks enforce modal alignment in Euclidean space while neglecting the manifold characteristics of cross-modal data, resulting in unnatural fused images with severe detail loss, and this often produces suboptimal fusion results. In contrast, our SPDAM adheres to the intrinsic topological structure between modalities through manifold space modeling, achieving a balance between detail preservation and semantic integrity.}
    \label{Ablation3}
\end{figure}

\subsubsection{Fusion Results on RoadScene}
Fig. \ref{RoadScene} presents a qualitative comparison on the RoadScene dataset. (c), (f), (g), (k) and (m) have lost some critical road marking information, which is quite crucial in the overall image. In the results obtained by (e), (i), (n) and (o), the pedestrians and ground appear too bright, which is not friendly to the human visual perception. (d), (h), (j) and (l) generally produced color distortion in the targets, leading to the loss of texture details. In contrast, the proposed method maximizes the retention of scene details while highlighting the intensity information of pedestrians in the infrared modality, achieving perfect fusion.

The average values of the seven metrics are shown in the Table \ref{Quantitave RoadScene}, where SMLNet has reached the highest level on four metrics, even demonstrating remarkable advantages on three of them (MI, SD, VIF). The aforementioned observations indicate that the proposed manifold attention model can skillfully handle various complex scenes, producing clearer and more natural fusion results.

\begin{figure*}
    \centering
    \small
    \includegraphics[width=1\linewidth]{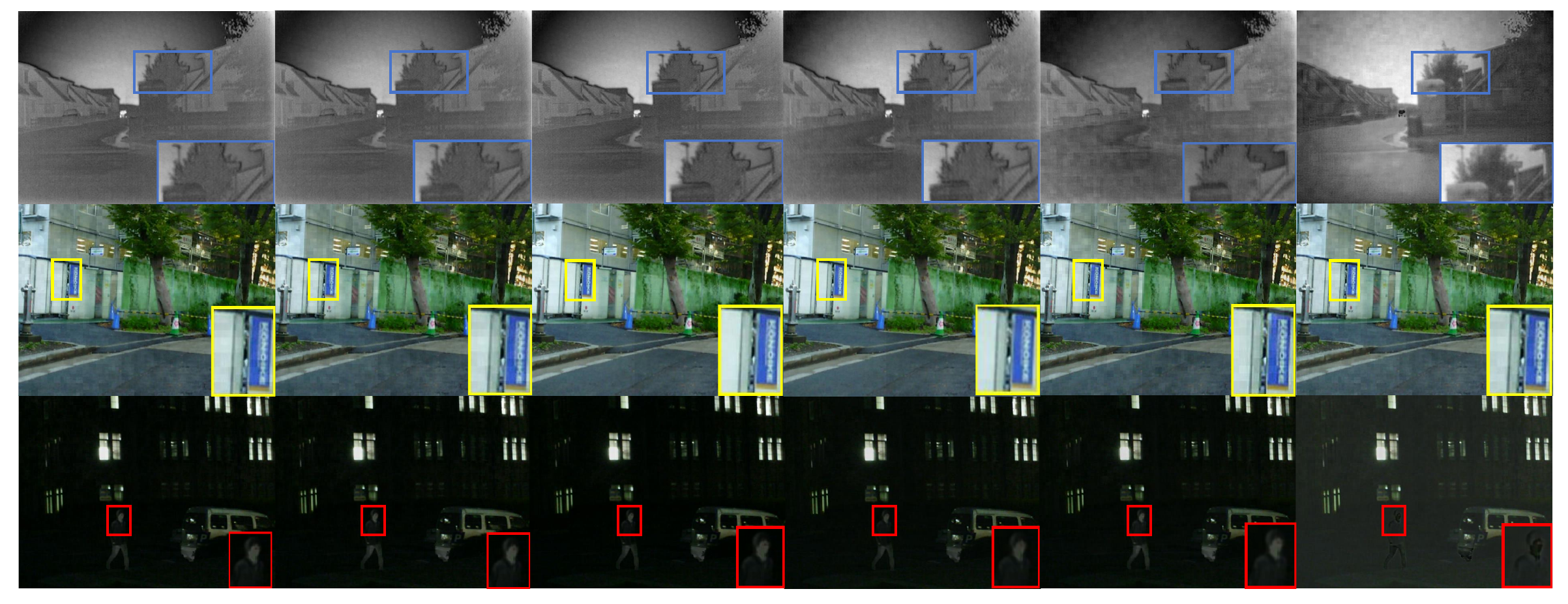}

    \begin{minipage}{0.16\textwidth}
        \centering \(\epsilon=0.01\)
    \end{minipage}
    \begin{minipage}{0.16\textwidth}
        \centering \(\epsilon=0.005\)
    \end{minipage}
    \begin{minipage}{0.16\textwidth}
        \centering \(\epsilon=0.001\)
    \end{minipage}
    \begin{minipage}{0.16\textwidth}
        \centering \(\epsilon=0.0005\)
    \end{minipage}
    \begin{minipage}{0.16\textwidth}
        \centering \(\epsilon=0.0001\)
    \end{minipage}
    \begin{minipage}{0.16\textwidth}
        \centering \mbox{w/o orth}
    \end{minipage}

    \caption{Visualization of the impact of different \(\epsilon\) values on the fusion results. Optimal SPD matrix regularization (\textit{e.g.}, \(1\mathrm{e}^{-3}\)) balances detail preservation and semantic integrity, while improper  values cause artifacts or loss of critical features. In addition, the BiMap layer’s orthogonality constraint is essential to maintain Riemannian geometry, as random initialization distorts manifold structure and degrades fusion quality.}
    \label{Ablation4}
\end{figure*}

\begin{table*}[ht!]
\centering % Centering the table
% \captionsetup{justification=justified,singlelinecheck=false}
\caption{Objective results of ablation study on TNO dataset. Here, ``b2-r2" and ``b3-r3" represent the number of blocks in BiMap and ReEig, respectively. ``single-modal strategy(ir)" and ``single-modal strategy(vi)" indicate the modal information contained within the SPD matrix that needs to be learned. ``\(F_{\text{Dense}}\)", ``\(F_{\text{SE}}\)", ``\(F_{\text{CBAM}}\)", ``\(F_{\text{ViT}}\)" denote that our SPDAM is replaced by Dense Block, SE Block, CBAM, and ViT structures, respectively. ``\(\epsilon\)" is a perturbation applied to the images. The highest result is highlighted in \textbf{\textsl{bolditalic}}.}

\label{ablation study}
% Table caption
\begin{tabular}{cccccc} % There are 8 centered columns (c), you can change to l (left align) or r (right align) as needed
\toprule
      & EN & SD & SF & VIF & \(Q^{AB/F}\)  \\
\midrule
 Ours(b1-r1) & 7.068 & 43.209 & \textbf{\textsl{11.098}} & \textbf{\textsl{0.832}} & \textbf{\textsl{0.516}} \\
\midrule
 b2-r2 & 7.000 & 39.857 & 9.993 & 0.638 & 0.449 \\
 b3-r3 & 6.991 & 39.412 & 9.784 & 0.633 & 0.443 \\
\midrule
 single-modal strategy(ir) & 7.018 & 41.481 & 9.049 & 0.616 & 0.426  \\ 
 single-modal strategy(vi) & 6.989 & 41.003 & 10.363 & 0.641 & 0.449 \\
 \midrule
 \(F_{\text{Dense}}\) & 6.902 & 35.517 & 7.322 & 0.579 & 0.395 \\
 \(F_{\text{SE}}\) & 7.102 & 46.090 & 6.337 & 0.564 & 0.302 \\
 \(F_{\text{CBAM}}\) & \textbf{\textsl{7.121}} & \textbf{\textsl{49.124}} & 7.625 & 0.673 & 0.378 \\
 \(F_{\text{ViT}}\) & 6.950 & 39.112 & 9.069 & 0.643 & 0.428 \\
\midrule
\(\epsilon\)=0.01 & 6.988 & 39.708 & 10.164 & 0.662 & 0.464 \\
\(\epsilon\)=0.0001 & 6.974 & 37.939 & 8.929 & 0.504 & 0.381 \\
\(\epsilon\)=0.005 & 7.001 & 39.863 & 10.054 & 0.652 & 0.456  \\
\(\epsilon\)=0.0005 & 7.003 & 40.429 & 9.600 & 0.564 & 0.401 \\\
w/o orth & 6.947 & 39.523 & 9.789 & 0.772 & 0.499 \\

\bottomrule
\end{tabular}
\end{table*}

\subsection{Ablation Study}
In this section, we objectively evaluate the effectiveness of key components in our framework through comparative analyses, including fusion strategies, SPD-block layer structures, the rationality of manifold attention, different manifold constraints, manifold distributions, and noise robustness tests.

\begin{figure*}
    \centering
    \small
    % \captionsetup{justification=justified,singlelinecheck=false}
    \includegraphics[width=1\linewidth]{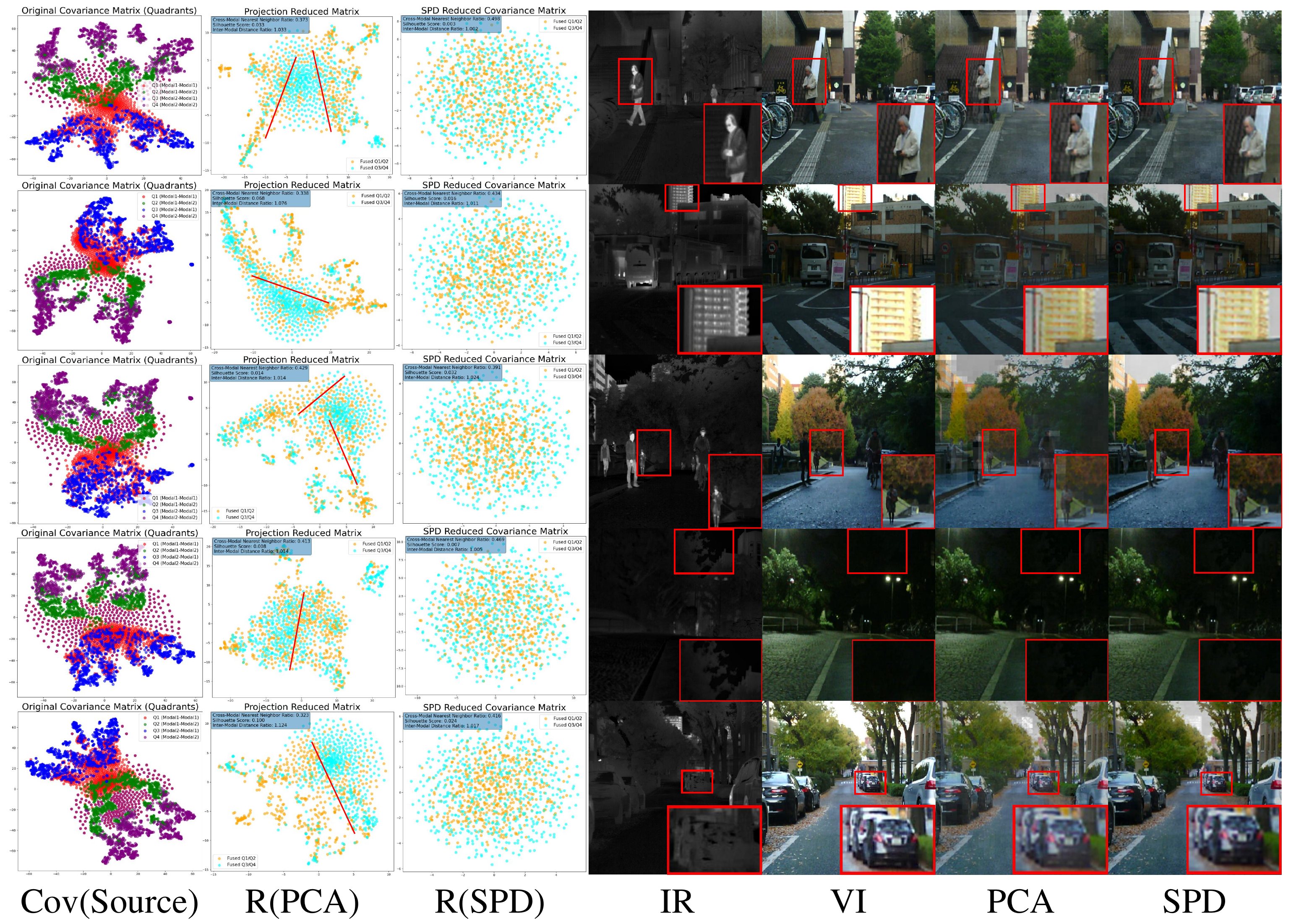}
    
    \caption{Covariance row vectors visualized via t-SNE under PCA versus SPD manifold learning dimensionality reduction, color-coded by quadrant and supplemented with fusion results. Traditional methods exhibit modality-biased clustering, manifesting as distinct separations between Fused Q1/Q2 (modality-1-dominant) and Q3/Q4 (modality-2-dominant) distributions, as demarcated by red guidelines. In contrast, our manifold-based framework establishes a unified statistical representation for cross-modal images, achieving homogeneous fusion of multi-modal statistics. This uniform distribution eliminates modality dominance artifacts while preserving thermally salient structures and textural details, directly enhancing fusion quality.}
    \label{Ablation5}
\end{figure*}

\subsubsection{Different Fusion Strategies}
In this section, we discussed in detail how different fusion strategies affect the fusion results and compared them from subjective and objective perspectives. ``cross-modal strategy" refers to a fusion strategy that computes the covariance matrices of two modalities, while ``single-modal strategy" refers to computing the covariance matrices of each modality individually. Table \ref{ablation study} presents the average values of five metrics. From a quantitative perspective, the cross-modal fusion strategy exhibits better performance in all five metrics: EN, SD, SF, VIF, and \(Q^{AB/F}\). This demonstrates that the design of our fusion strategy is beneficial in making the fused images visually closer to real scenes, effectively preserving the rich texture details of the source images. Some examples of the fusion results obtained using different fusion strategies are shown in Fig. \ref{Ablation1}. When only the infrared (IR) modality is used, the results lose texture details from visible images. Conversely, relying solely on the visible (VI) modality causes the loss of target brightness information, which is unacceptable for image fusion tasks. Therefore, in our framework, we adopt a cross-modal fusion strategy to integrate information from multiple source images and generate satisfactory fused images.

\begin{table}
\centering % Centering the table
\small
% \captionsetup{justification=justified,singlelinecheck=false}
\caption{The average values of different manifold distribution metrics on the MSRS dataset. The top two metrics are highlighted in \textbf{\textsl{bolditalic}} and \textbf{bold}.}

\label{manifold distribution}
% Table caption
\begin{tabular}{ p{1.6cm} p{1.4cm} p{1.2cm} p{1.2cm}  } % There are 8 centered columns (c), you can change to l (left align) or r (right align) as needed
\toprule
     & CM-NNR & SS & IMDR  \\
\midrule
 PCA & \textbf{0.375} & \textbf{0.045} & \textbf{1.052} \\
 SPD(Ours) & \textbf{\textsl{0.442}} & \textbf{\textsl{0.016}} & \textbf{\textsl{1.012}} \\
\bottomrule
\end{tabular}
\end{table}

\subsubsection{Different Layer Structures of SPD-Block}
When processing data on Riemannian manifolds, each layer of BiMap and ReEig signifies a transformation of the Riemannian manifold. The processing of each layer needs to maintain the topological characteristics of the data, that is, to preserve the proximity and the interrelations between data points. If the structural properties of the original Riemannian manifold are lost during the multi-layer processing, the quality of the final fusion will decrease. 

Deep architectures may not always exhibit optimal performance in low-level visual tasks such as image fusion. Therefore, we attempted to train using different manifold network structures. As shown in Table \ref{ablation study} and Fig. \ref{Ablation2}, when the number of layers of BiMap and ReEig is only set to 1, the features of the original data are faithfully preserved through single-layer transformation, reducing the distortion or information loss that may occur in the features after multi-layer mapping. Meanwhile, as the number of layers in the SPD network increases, the gradient update may bias towards the deep semantic features within the modality due to their larger variance, while fusion tasks often require more attention to pixel-level information. Therefore, in our work, we adopt one layer of BiMap and ReEig to achieve the best performance.

\begin{figure}
    \small
    \centering
    % \captionsetup{justification=justified,singlelinecheck=false}
    \includegraphics[width=1\linewidth]{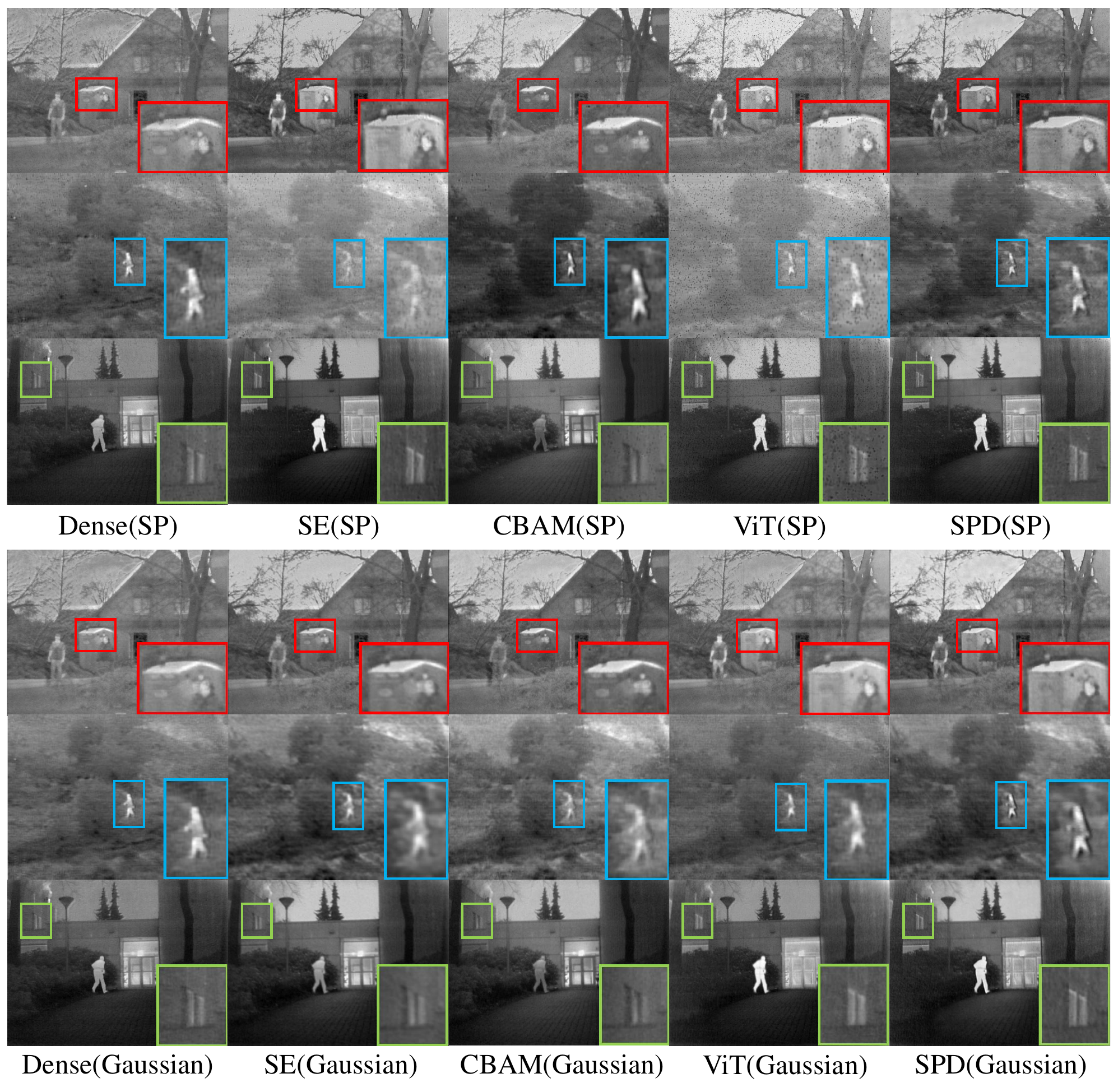}
    
    \caption{Euclidean networks employ linear operations (\textit{e.g.}, convolution, matrix multiplication) and demonstrate heightened sensitivity to both Gaussian and salt-and-pepper noise due to their lack of geometric constraints, resulting in significant feature map degradation. In contrast, manifold networks effectively suppress high-amplitude noise interference through nonlinear geometric mappings while preserving the structural integrity of critical features.}
    \label{Ablation6}
\end{figure}

\subsubsection{Manifold Attention}
To evaluate the effectiveness of SPDAM, we conduct experiments by replacing it with four representative network architectures, including both classic designs (\textit{e.g.}, DenseNet) and attention-based structures (\textit{e.g.}, SENet, CBAM, Transformer). For a more intuitive comparison, we visualize both the source images and the fused images from different methods. As shown in Fig. \ref{Ablation3}, compared to previous traditional methods, our SMLNet enhances the color and brightness characteristics and also highlights the detail features of the pedestrian area in the infrared modality. This shows that the manifold attention follows the intrinsic topological structure between modalities, making the input end semantically tend to be balanced.

At the same time, we also conducted a quantitative comparative experiment to further demonstrate the effectiveness of our SPDAM in the fusion task. As shown in Table \ref{ablation study}, our module surpasses other methods in terms of SF, VIF and \(Q^{AB/F}\), indicating that under the iteration of the Riemannian network, the network has deepened its focus on the associated information of the salient parts within modalities. Moreover, it emphasizes the useful complementary information between different modalities, optimizing fused results for enhanced contrast while preserving structural detail.

\begin{table}
\centering % Centering the table
\small
% \captionsetup{justification=justified,singlelinecheck=false}
\caption{The average quantitative results of different network noise additions on the TNO dataset. The top two metrics are highlighted in \textbf{\textsl{bolditalic}} and \textbf{bold}.}

\label{noise}
% Table caption
\begin{tabular}{ p{1.9cm} p{0.7cm} p{0.7cm} p{0.7cm} p{0.7cm} p{0.8cm}  } % There are 8 centered columns (c), you can change to l (left align) or r (right align) as needed
\toprule
   Gaussian  & EN & SD & SF & VIF & \(Q^{AB/F}\) \\
\midrule
 \(F_{\text{Dense}}\) & 6.943 & 36.238 & 7.404 & 0.459 & 0.284  \\
 \(F_{\text{SE}}\) & \textbf{7.048} & \textbf{40.969} & 5.453 & 0.453 & 0.240\\
 \(F_{\text{CBAM}}\) & 7.033 & 40.357 & 6.837 & \textbf{\textsl{0.472}} & 0.274 \\
 \(F_{\text{ViT}}\) & 6.992 & 39.171 & \textbf{7.512} & 0.459 & \textbf{\textsl{0.295}} \\
SMLNet(Ours) & \textbf{\textsl{7.070}} & \textbf{\textsl{42.199}} & \textbf{\textsl{7.830}} & \textbf{0.469} & \textbf{0.292} \\
\midrule
   SP & EN & SD & SF & VIF & \(Q^{AB/F}\) \\
\midrule
 \(F_{\text{Dense}}\) & 6.899 & 34.950 & 8.925 & 0.432 & 0.310  \\
 \(F_{\text{SE}}\) & \textbf{7.022} & 40.514 & 7.692 & \textbf{\textsl{0.498}} & 0.308\\
 \(F_{\text{CBAM}}\) & 6.989 & \textbf{40.614} & 9.739 & 0.443 & 0.296 \\
 \(F_{\text{ViT}}\) & 7.001 & 39.029 & \textbf{\textsl{16.658}} & 0.331 & \textbf{0.343}\\
SMLNet(Ours) & \textbf{\textsl{7.044}} & \textbf{\textsl{40.736}} & \textbf{11.482} & \textbf{0.487} & \textbf{\textsl{0.369}} \\
\bottomrule
\end{tabular}
\end{table}

\subsubsection{The Impact of Different Manifold Constraints}
Equation \ref{EIG} enhances the discriminative of SPD matrices through the correction of small eigenvalues. Since the shortest path between points on a Riemannian manifold is represented by geodesic distance, the epsilon setting in the ReEig layer is effectively adjusting the curvature on the manifold, thereby affecting the representation of local contrast and textural information in images. As shown in Fig. \ref{Ablation4}, we selected an instance from the MSRS dataset for intuitive demonstration. When \(\epsilon\) is set to \(1\mathrm{e}^{-2}\) and \(1\mathrm{e}^{-4}\), the image exhibits obvious block artifacts due to uneven SPD weight distribution, which affects the visual quality of the image. When \(\epsilon\) is set to \(5\mathrm{e}^{-3}\) and \(5\mathrm{e}^{-4}\), the image loses edge details and shows diminished object contours. When \(\epsilon\) is set to \(1\mathrm{e}^{-3}\), the fused image achieves the optimal effect.

Meanwhile, as shown in Table \ref{ablation study}, when \(\epsilon\) is too large, the excessive diagonal perturbation oversmooths the covariance structure, suppressing subtle details and high-frequency components. This reduces EN and SD, as the fused image loses nuanced variations. Conversely, an overly small \(\epsilon\) inadequately regularizes the covariance matrix, weakening the emphasis on cross-modal edge and texture correlations. This results in diminished SF, VIF, and \(Q^{AB/F}\), as critical inter-modal features are insufficiently integrated. Therefore, by selecting an appropriate perturbation value, we achieve an optimal balance between preserving high-frequency image details and effectively regularizing the covariance matrix.

Moreover, the BiMap layer, by design, enforces orthogonality or near-orthogonality conditions on its weight matrices to ensure that transformations respect the underlying Riemannian geometry of the data. This is critical for maintaining the discriminative properties of the features, because it preserves angles, distances, and relative magnitudes during linear transformations.

When remove this constraint and initialize the matrix randomly, the transformation no longer adheres to the Riemannian structure. The random matrix may introduce arbitrary scaling, skewing, or even collapsing of the feature space. For example, the white highlights might collapse to lower values because the transformation no longer preserves the relative ordering. As shown in Fig. \ref{Ablation4}, a ``distortion" occurs on the manifold, disrupting the semantic and geometric relationships that the network was originally designed to encode.

The observed results align with our theoretical expectation that removing the orthogonality constraint would lead to severe image degradation.

\begin{figure*}
    
    \centering
    % \captionsetup{justification=justified,singlelinecheck=false}
    \includegraphics[width=\linewidth]{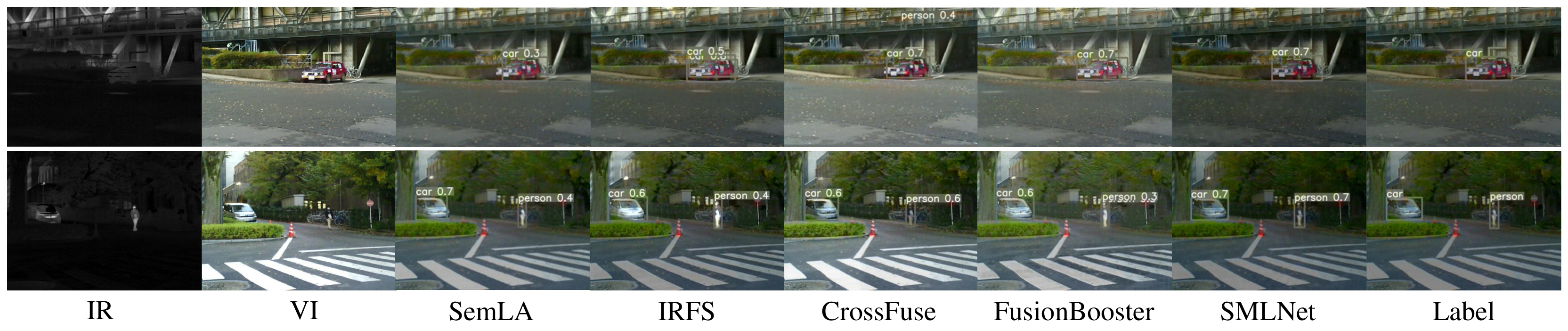}
    
    % \begin{minipage}{0.15\textwidth}
    %     % \centering
    %     IR
    % \end{minipage}
    % % \hspace{0.5pt} % 文本之间的空格
    % \begin{minipage}{0.09\textwidth}
    %     \centering
    %     VI
    % \end{minipage}
    % % \hspace{0.5pt} % 文本之间的空格
    % \begin{minipage}{0.09\textwidth}
    %     % \centering
    %     IR
    % \end{minipage}
    % % \hspace{0.5pt} % 文本之间的空格
    % \begin{minipage}{0.09\textwidth}
    %     % \centering
    %     VI
    % \end{minipage}
    % % \hspace{0.5pt} % 文本之间的空格
    % \begin{minipage}{0.09\textwidth}
    %     % \centering
    %     DeFusion
    % \end{minipage}
    % \begin{minipage}{0.09\textwidth}
    %     % \centering
    %     SemLA
    % \end{minipage}
    % \begin{minipage}{0.09\textwidth}
    %     % \centering
    %     CrossFuse
    % \end{minipage}
    % \begin{minipage}{0.09\textwidth}
    %     % \centering
    %     \small
    %     FusionBooster
    % \end{minipage}
    % \begin{minipage}{0.09\textwidth}
    %     % \centering
    %     SMLNet
    % \end{minipage}
    % \begin{minipage}{0.09\textwidth}
    %     % \centering
    %     Label
    % \end{minipage}

    \hspace{1pt} % 文本之间的空格

    \caption{Object detection results compared with other state-of-the-art fusion methods. Unlike traditional methods that simply concatenate multi-modal features, SMLNet's manifold fusion mechanism adaptively learns shared feature representations of key targets (\textit{e.g.}, edges and textures). This approach notably enhances the detector's sensitivity to small and blurred objects while effectively reducing false positives and missed detections.}
    \label{detection}
\end{figure*}

\begin{figure*}
    
    \centering
    % \captionsetup{justification=justified,singlelinecheck=false}
    \includegraphics[width=\linewidth]{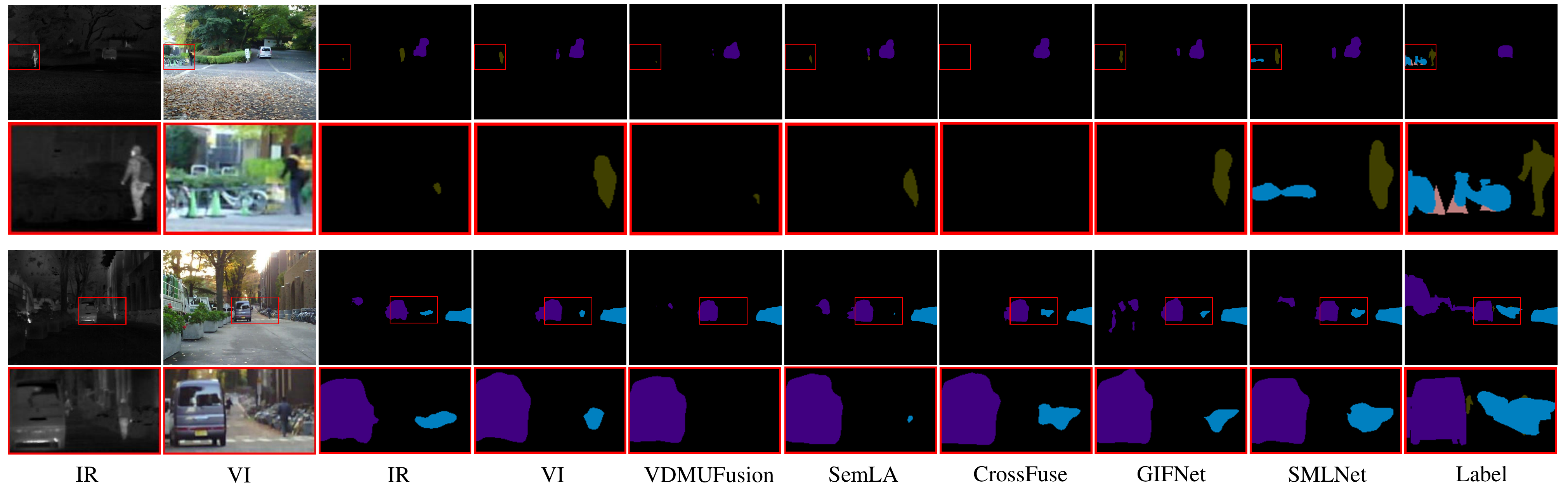}
    
    % \begin{minipage}{0.15\textwidth}
    %     % \centering
    %     IR
    % \end{minipage}
    % % \hspace{0.5pt} % 文本之间的空格
    % \begin{minipage}{0.09\textwidth}
    %     \centering
    %     VI
    % \end{minipage}
    % % \hspace{0.5pt} % 文本之间的空格
    % \begin{minipage}{0.09\textwidth}
    %     % \centering
    %     IR
    % \end{minipage}
    % % \hspace{0.5pt} % 文本之间的空格
    % \begin{minipage}{0.09\textwidth}
    %     % \centering
    %     VI
    % \end{minipage}
    % % \hspace{0.5pt} % 文本之间的空格
    % \begin{minipage}{0.09\textwidth}
    %     % \centering
    %     DeFusion
    % \end{minipage}
    % \begin{minipage}{0.09\textwidth}
    %     % \centering
    %     SemLA
    % \end{minipage}
    % \begin{minipage}{0.09\textwidth}
    %     % \centering
    %     CrossFuse
    % \end{minipage}
    % \begin{minipage}{0.09\textwidth}
    %     % \centering
    %     \small
    %     FusionBooster
    % \end{minipage}
    % \begin{minipage}{0.09\textwidth}
    %     % \centering
    %     SMLNet
    % \end{minipage}
    % \begin{minipage}{0.09\textwidth}
    %     % \centering
    %     Label
    % \end{minipage}

    \hspace{1pt} % 文本之间的空格

    \caption{Segmentation results compared with other state-of-the-art fusion methods. Existing methods commonly suffer from semantic information loss and misclassification issues. In contrast, SMLNet employs a correlation-based semantic weighting mechanism to enhance the contrast of discriminative regions while effectively suppressing irrelevant background interference.}
    \label{segmentation}
\end{figure*}

\begin{table*}[ht]
\centering % Centering the table
% \captionsetup{justification=justified,singlelinecheck=false}
\caption{Quantitative evaluation of object detection using the MSRS dataset. The best three metrics are highlighted in \textbf{\textsl{bolditalic}}, \textbf{bold} and \textsl{italic} fonts, respectively.}    % Table caption
\label{Quantitative detection}
\begin{tabular}{cccccc} % There are 8 centered columns (c), you can change to l (left align) or r (right align) as needed
\toprule
Methods & Person & Car & All & mAP@0.5 & mAP@0.5:0.95  \\
\midrule
DenseFuse~\citep{li2018densefuse} & 0.730 & 0.642 & 0.686 & 0.792 & 0.423\\
RFN-Nest~\citep{11} & 0.619 & 0.611 & 0.615 & 0.750 & \textsl{0.482} \\
DeFusion~\citep{liang2022fusion} & \textbf{0.937} & 0.798 & \textbf{0.868} & 0.805 & 0.453 \\
SemLA~\citep{xie2023semantics} & 0.775 & 0.646 & 0.711 & 0.767 & 0.519 \\
DDFM~\citep{zhao2023ddfm} & \textsl{0.932} & \textsl{0.801} & \textsl{0.867} & \textsl{0.810} & \textbf{0.483}\\
LRRNet~\citep{8} & 0.577 & \textbf{0.842} & 0.709 & 0.721 & 0.404\\
IRFS~\citep{wang2023interactively} & 0.773 & 0.799 & 0.786 & \textbf{0.819} & \textsl{0.482} \\
CrossFuse~\citep{li2024crossfuse} & 0.926 & 0.767 & 0.846 & 0.803 & 0.473\\
VDMUFusion~\citep{shi2024vdmufusion} & 0.750 & 0.514 & 0.632 & 0.760 & 0.439 \\
DCINN~\citep{2024dcinn} & 0.611 & 0.622 & 0.617 & 0.761 & 0.456 \\
DDBFusion~\citep{zhang2025ddbfusion} & 0.712 & 0.710 & 0.711 & 0.771 &  0.453 \\
FusionBooster~\citep{cheng2024fusionbooster} & 0.619 & 0.704 & 0.661 & 0.774 & 0.422\\
GIFNet~\citep{cheng2025cvpr_gifnet} & 0.512 & 0.740 & 0.626 & 0.693 & 0.388 \\
\midrule
SMLNet & \textbf{\textsl{0.939}} & \textbf{\textsl{0.997}} & \textbf{\textsl{0.968}} & \textbf{\textsl{0.822}} & \textbf{\textsl{0.545}} \\
\bottomrule
\end{tabular}
\end{table*}

\begin{table*}[ht!]
\centering % Centering the table
% \captionsetup{justification=justified,singlelinecheck=false}
\caption{Quantitative evaluation of segmentation using the MSRS dataset. The best three metrics are highlighted in \textbf{\textsl{bolditalic}}, \textbf{bold} and \textsl{italic} fonts, respectively.}    % Table caption
\label{Quantitative segmentation}
\begin{tabular}{cccccc} % There are 8 centered columns (c), you can change to l (left align) or r (right align) as needed
\toprule
Methods & Unl & Car & Per & Bik & mIOU  \\
\midrule
VI & \textsl{97.89} & \textsl{74.75} & 35.31 & 62.59 & 67.63 \\
IR & 96.93 & 61.13 & 15.14 & 45.77 & 54.74 \\
DenseFuse~\citep{li2018densefuse} & 97.82 & 73.91 & 24.70 & \textbf{\textsl{65.37}} & 65.45 \\
RFN-Nest~\citep{11} & \textbf{\textsl{97.91}} & \textbf{\textsl{75.35}} & 31.90 & 61.81 & 66.74 \\
DeFusion~\citep{liang2022fusion} & 97.85 & 74.18 & 21.84 & 64.70 & 64.64 \\
SemLA~\citep{xie2023semantics} & 97.84 & 73.87 & 32.96 & 59.95 & 66.16 \\
DDFM~\citep{zhao2023ddfm}& 97.85 & 72.90 & 33.37 & 61.33 & 66.36 \\
LRRNet~\citep{8} & 97.86 & 73.57 & \textbf{\textsl{40.06}} & 59.21 & \textsl{67.68} \\
IRFS~\citep{wang2023interactively} & 97.87 & 73.41 & \textbf{\textsl{42.12}} & 58.13 & \textbf{\textsl{67.88}} \\
CrossFuse~\citep{li2024crossfuse} & 97.86 & 73.70 & 33.00 & 62.07 & 66.66 \\
VDMUFusion~\citep{shi2024vdmufusion} & 97.80 & 73.10 & 32.92 & 59.07 & 65.73 \\
DCINN~\citep{2024dcinn} & 97.78 & 72.42 & 35.98 & 57.60 & 65.94 \\
DDBFusion~\citep{zhang2025ddbfusion} & 97.85 & 73.06 & 33.22 & 63.15 & 66.82 \\

FusionBooster~\citep{cheng2024fusionbooster} & 97.87 & 72.78 & 34.38 & \textsl{65.30} & 67.58 \\
GIFNet~\citep{cheng2025cvpr_gifnet} & 97.85 & 73.19 & 35.98 & 61.30 & 67.08 \\
\midrule
SMLNet & \textbf{\textsl{98.19}} & \textbf{\textsl{78.87}} & \textsl{37.19} & \textbf{\textsl{66.13}} & \textbf{\textsl{70.09}} \\
\bottomrule
\end{tabular}
\end{table*}

\subsubsection{Comparison of Distributions under Different Learning Paradigms}
In image fusion, effective fusion requires preserving complementary cross-modal interactions while respecting the intrinsic Riemannian geometry of the source data. To validate the geometric advantages of the manifold learning in SMLNet, we visualize original and dimensionality-reduced covariance matrices from MSRS samples using t-SNE. As shown in Fig. \ref{Ablation5}, each point corresponds to a row vector of the original covariance matrix, with Q1-Q4 representing statistical correlation blocks from four quadrants. To explicitly decouple modality-specific statistical signatures, we partition the covariance components into two geometrically meaningful groups: Q1/Q2 capturing modality-1-dominated covariance structures (infrared self-characteristics with cross-modal correlations) and Q3/Q4 representing modality-2-dominated covariance geometry (visible self-characteristics with complementary interactions). The t-SNE visualization result reveals: Traditional PCA forces linear projections onto Euclidean axes, flattening the curved SPD manifold structure and catastrophically compressing both modality-specific features (Q1/Q4) and cross-modal interactions (Q2/Q3). This distortion induces regional artifacts and thermal-textural misalignment (Rows 1 
\&
3). Conversely, our SPD transformations (BiMap layers) preserve intrinsic manifold geometry. Critically, t-SNE visualization confirms that our method uniformly blends modality-1-dominated and modality-2-dominated feature clusters on the SPD manifold. By performing dynamic Riemannian weighting within this geometrically consistent space, we optimally fuse complementary information while resolving inter-modal conflicts. For instance, in scenes like streetlights (Row 2), foliage (Row 4), and vehicles (Row 5), infrared thermal variations and visible texture gradients are jointly encoded through SPD-manifold-based cross-modal interactions. The result is fused imagery preserving thermal contrast while enhancing structural edges, well-aligned with human vision.

To quantitatively assess the fusion quality, we employ the following metrics: Silhouette Score (SS), Inter-Modal Distance Ratio (IMDR) and Cross-Modal Nearest Neighbor Ratio (CM-NNR). The SS is defined as:
\begin{equation}
    \text{SS} = \frac{1}{N} \sum_{i=1}^{N} \frac{b(i) - a(i)}{\max\{a(i), b(i)\}}
    ,
\end{equation}
where \(a(i)\) is the average distance between sample \(i\) and all other samples within the same modality, measuring intra-modal cohesion. \(b(i)\) denotes the smallest average distance from sample \(i\) to samples in any other modality, reflecting inter-modal separation.

IMDR evaluates fusion uniformity by dividing the mean inter-modal distance by the mean intra-modal distance:
\begin{equation}
    \text{IMDR} = \frac{D_{\text{inter}}}{D_{\text{intra}}},
\end{equation}
where \(D_{\text{inter}}\) is the mean pairwise distance between samples from different modalities:
\begin{equation}
    D_{\text{inter}} = \frac{1}{|S_{\text{inter}}|}\sum_{(i,j)\in S_{\text{inter}}} d(x_i, x_j),
\end{equation}
where \(D_{\text{intra}}\) is the mean pairwise distance between samples within the same modality:
\begin{equation}
    D_{\text{intra}} = \frac{1}{|S_{\text{intra}}|}\sum_{(i,j)\in S_{\text{intra}}} d(x_i, x_j)
    ,
\end{equation}
where \(S_{\text{inter}}\) denotes the set of all cross-modal sample pairs and \(S_{\text{intra}}\) represents the set of all intra-modal sample pairs. \(d(x_i, x_j)\) is the distance metric (\textit{e.g.}, Euclidean distance) between samples \(x_{i}\) and \(x_{j}\).

CM-NNR quantifies mixing by averaging the fraction of neighbors from another modality for each point. For each sample \(i\), the CM-NNR is calculated as:
\begin{equation}
    \text{CM-NNR}(i) = \frac{|\{j \in \text{NN}_k(i) | \text{label}(j) \neq \text{label}(i)\}|}{k}
    ,
\end{equation}
where \(\text{NN}_k(i)\) is the set of k-nearest neighbors of sample \(i\), \(\text{label}(j)\) denotes the modality label of sample \(j\), \(k\) represents the number of nearest neighbors considered.

The global CM-NNR is computed as:
\begin{equation}
    \text{CM-NNR} = \frac{1}{N}\sum_{i=1}^{N} \text{CM-NNR}(i).
\end{equation}

As shown in Table \ref{manifold distribution}, our SPD-manifold-based data points achieved the highest scores across all three metrics. The superior SS scores confirm enhanced feature compactness within modalities while maintaining clear separation between them. Optimal IMDR values demonstrate our method's unique ability to balance cross-modal distances, while the highest CM-NNR indicates thorough modality mixing. These quantitative advantages directly validate that our SPD-manifold framework effectively eliminates heterogeneity while achieving tighter, more balanced fusion.

\subsubsection{Noise Robustness in Riemannian vs Euclidean Networks}
The presence of noise in real scene data is known to potentially distort feature representations and their statistical relationships, the robustness of multi-modal fusion methods heavily depends on their ability to maintain stable feature representations under noisy conditions. To further investigate this phenomenon, we conducted additional experiments.

At the input stage, we introduced controlled noise injection to artificially perturb the multi-modal data. As shown in Fig. \ref{Ablation6}, we systematically compared the fusion performance differences under varying noise levels. While Euclidean-based networks (especially ViT) demonstrate strong performance in low-noise scenarios due to the long-range modeling capability of attention mechanisms, they exhibit heightened sensitivity to cross-modal disturbances (\textit{e.g.}, discrete impulses from salt-and-pepper noise and continuous contamination from Gaussian noise), resulting in residual noise artifacts and blurring effects. In contrast, manifold-based networks model local statistical properties via covariance matrices. Since covariance captures second-order statistical relationships rather than raw pixel values, outliers are naturally suppressed during computation, enabling better preservation of edge structures and cross-modal features during denoising.

In Table \ref{noise}, our manifold network delivers exceptional results in Gaussian noise removal, achieving near-state-of-the-art performance on both VIF and \(Q^{AB/F}\) metrics. A noteworthy observation is that in salt-and-pepper noise scenarios, the SF metric exhibits non-linear correlation with fusion quality, attributable to the distinctive interference patterns of synthetic noise. Remarkably, our method achieves the optimal trade-off between information preservation and noise suppression, demonstrating the inherent stability of manifold-based feature representation.

\subsection{Experiments in Object Detection}

\subsubsection{Quantitative Comparisons}
We performed an evaluation of the previously mentioned fusion networks using YOLOv7 ~\citep{wang2023yolov7}. The results were measured by the object detection metrics, namely: accuracy, mean precision at 0.5 (AP50) and mean precision at 0.5:0.95 (AP50:95). As shown in Table \ref{Quantitative detection}, our method achieved the highest performance on all five metrics.

\subsubsection{Qualitative Comparisons}
In terms of visual effects, our method also has significant advantages, as shown in Fig. \ref{detection}. In the examples of tradition methods, SemLA failed to detect the car, while IRFS exhibited cases of redundant detection without accurately distinguishing the targets. Although CrossFuse have done well in the above aspects, false positives in regions with unclear edges have led to less than ideal detection results. In addition, FusionBooster has relatively limited capability in pedestrian detection. In contrast, our SMLNet maintained high detection accuracy in these challenging scenarios, preserving the prominent features and texture details of the targets, which further verifies the superiority of our approach.

\subsection{Experiments in Semantic Segmentation}

\subsubsection{Quantitative Comparisons}
To further validate the performance of our method on downstream vision tasks, we have conducted experiments on the MSRS segmentation dataset. During training, we used DeeplabV3+~\citep{chen2018encoder} for network training, with the SGD optimizer and cross-entropy loss function for parameter updates. As shown in Table \ref{Quantitative segmentation}, we comprehensively measured the performance using the average accuracy and mIOU (Mean Intersection over Union) for 4 pixel-level category labels (car, person, bike and background).

SMLNet achieved the best segmentation effect on the background, car and bike categories, and obtained the highest mIOU. At the same time, the Person category also reached the 3rd level. This indicates that the manifold attention guides the intra-modal strongly correlated features to perform weight redistribution, ensuring the semantic consistency within the original image.
\subsubsection{Qualitative Comparisons}
Fig. \ref{segmentation} depicts a visual comparison between SMLNet and other competing methods. As demonstrated in the first two example rows, our method effectively highlights salient information from the infrared modality in challenging scenarios, while other methods exhibit severe misclassification. It is also noteworthy that, in daytime scenarios, the shape and position information of objects that are closely matched with the environmental colors can be accurately estimated, as seen in the last two rows of examples with the protrusion at the rear of the bicycles and the roof of the cars. In summary, these two typical examples vividly demonstrate the advantages of our method over other competitors.

\begin{table}[t]
\centering % Centering the table
\small
% \captionsetup{justification=justified,singlelinecheck=false}
\caption{Efficiency comparison between SMLNet and 13 SOTA methods. The best three metrics are highlighted in \textbf{\textsl{bolditalic}}, \textbf{bold} and \textsl{italic} fonts, respectively.}

\label{efficiency comparison}
% Table caption
\begin{tabular}{ p{3cm} p{2.2cm} p{2.2cm}} % There are 8 centered columns (c), you can change to l (left align) or r (right align) as needed
\toprule

   Methods  & Params(MB) & FLOPs(G) \\
\midrule
DenseFuse & \textsl{0.074} & 159.170\\
RFN-Nest & 2.733 & 89.797 \\
DeFusion & 7.874 & \textsl{15.265} \\
SemLA & 7.278 & 16.613 \\
DDFM & 527.240 & 852.710 \\
LRRNet & \textbf{\textsl{0.049}} & \textbf{14.167} \\
IRFS & 0.242 & 78.655 \\
CrossFuse & 19.212 & 41.516 \\
VDMUFusion & 31.751 & 321.972 \\
DCINN & 2.097 & 17.026 \\
DDBFusion & 3.674 & 184.931 \\
FusionBooster & 0.555 & 56.407 \\
GIFNet & 0.613 & 39.814 \\
\midrule
SMLNet & \textbf{0.051} & \textbf{\textsl{3.652}} \\
\bottomrule
\end{tabular}
\end{table}

\subsection{Efficiency Comparison}
In this section, we present the model complexity and computational cost of comparative methods. As shown in the Table \ref{efficiency comparison}, methods such as SemLA, CrossFuse, IRFS, and GIFNet adopt the classic ViT architecture, resulting in high computational costs. DDBFusion, DeFusion, and DCINN employ relatively complex decomposition modules, while DDFM and VDMUFusion rely on generative models, introducing significant computational overhead. DenseFuse, FusionBooster, LRRNet, and RFN-Nest utilize simpler network structures, offering advantages in computational efficiency. In contrast, our SMLNet, due to its lightweight network architecture and manifold dimensionality reduction design, demonstrates significant advantages compared to existing methods.

\begin{figure}
    \small
    \centering
    % \captionsetup{justification=justified,singlelinecheck=false}
    \includegraphics[width=1\linewidth]{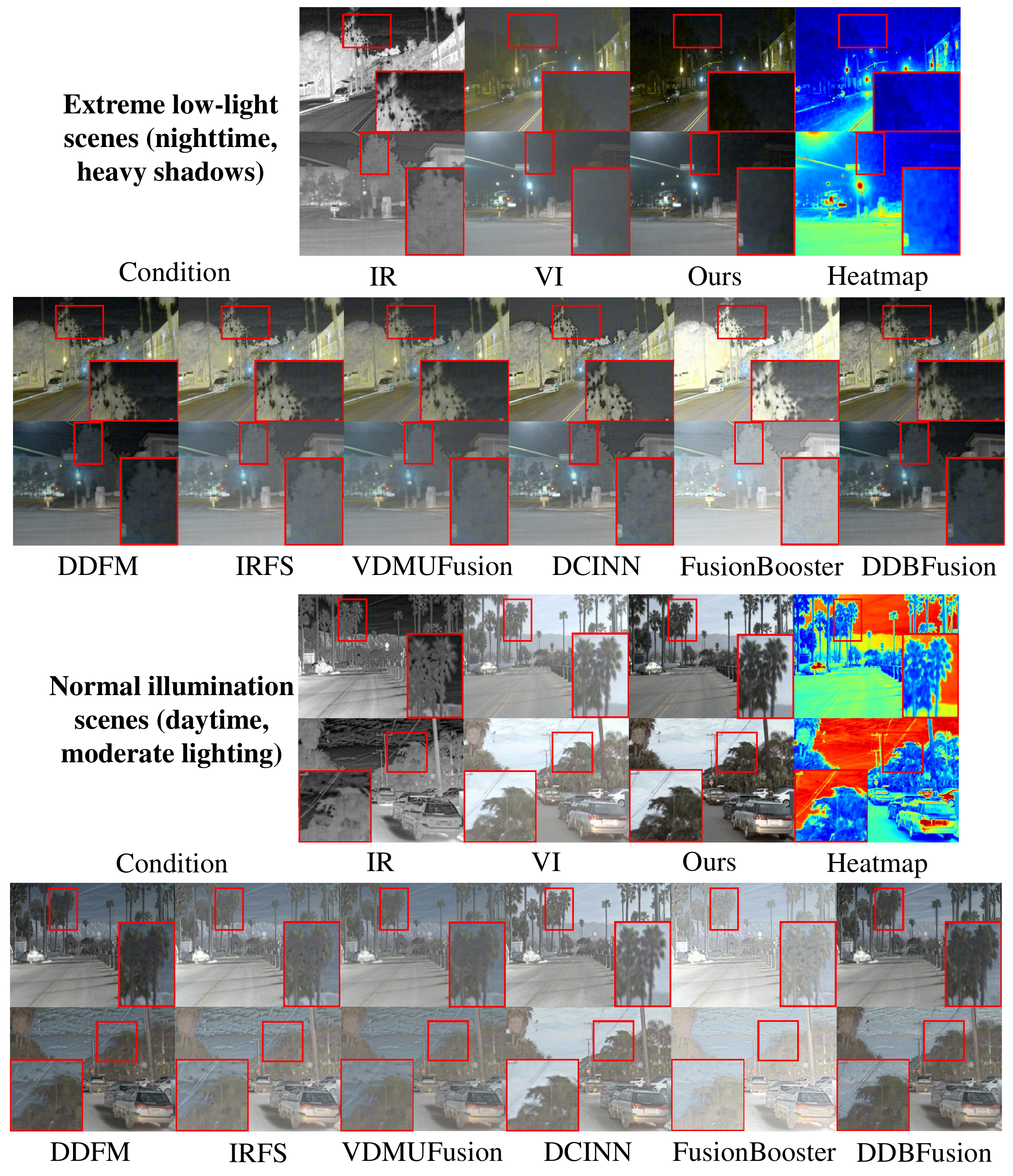}
    
    \caption{Comparative results of our fusion method under normal and extreme low-light conditions. The first and third rows display source images and their corresponding fusion results, while the second and fourth rows provide comparisons with other methods. In extreme low-light scenarios, the fusion quality is affected by error propagation from ill-conditioned covariance, leading to reduced contrast and subtle artifacts in dark regions. Under normal illumination, SMLNet effectively preserves critical details while integrating complementary cross-modal information.}
    \label{Condition}
\end{figure}

\subsection{Limitations in Extreme Scenarios Fusion}
Though SMLNet demonstrates robust performance across most multi-modal fusion tasks, its effectiveness may degrade under extreme low-light conditions, such as nighttime scenes in the RoadScene dataset, while alternative methods maintain basic functionality. As shown in Fig. \ref{Condition}, in these challenging scenarios, our method occasionally struggles to clearly distinguish between tree and sky regions, while also exhibiting block artifacts and amplified noise in dark areas.

In such regions, critically weak signals cause local patches in both infrared and visible modalities to exhibit near-uniform intensity values. This lack of variation drives the eigenvalues of the computed covariance matrix \(\mathbf{C}_{\mathbf{M}_{\text{ir}}\mathbf{M}_{\text{vi}}}\) toward zero, rendering it numerically singular. Although we employ SVD decomposition with a small perturbation term (Equation \ref{Epsilon}) to ensure positive definiteness, the extremely small eigenvalues still introduce numerical instability, affecting the subsequent dimensionality reduction in the BiMap layer and the reliability of SPD attention weights.

This issue is not a fundamental flaw of the method itself but arises because covariance matrix estimation relies on signal energy distribution. Under extremely low signal-to-noise ratio (SNR) conditions, noise dominates inter-modal relationships, degrading the accuracy of covariance-based feature mapping. In this scenario, global modeling approaches using second-order statistics show limited effectiveness, while methods relying on other principles may demonstrate relatively better performance. Consequently, future improvements should focus on enhancing covariance estimation robustness, optimizing eigenvalue conditioning, or exploring manifold-based fusion mechanisms adaptive to extreme SNR conditions.

\section{Conclusion}
Previous image fusion methods have primarily focused on computational processes that rely on Euclidean metric feature representations. Consequently, the intrinsic Riemannian of images have been ignored, resulting in the loss of some discriminative information when combining data from different modalities. 

In this paper, a pioneering approach is taken by modeling images onto the Riemannian SPD manifold and designing a non-Euclidean statistical correlation coding strategy. In this scenario, the source images are divided into small tokens and projected into the SPD network for manifold learning. This approach not only captures salient semantic information contained within each modality more effectively, but also enhances the model's overall perception of deep statistical information association between modalities.

Our attention module reformulates convolutional operations by treating the SPD matrix as a learnable weight kernel in eigenspace. This enables dynamic, geometry-aware feature modulation across pixel neighborhoods, directly responding to the intrinsic relationships within multi-modal data. By explicitly preserving the Riemannian structure during feature processing, our approach maintains critical high-frequency details and structural consistency in fused outputs, demonstrating superiority over Euclidean-space approaches.

Benchmark results on public datasets demonstrate that our approach not only retains richer visible-light details and enhances salient infrared features compared to existing fusion networks, but also achieves this with higher computational efficiency. Our fusion method has been applied to other computer vision tasks, such as object detection, and exhibited superior performance compared to other popular approaches as well.

In this work, we have only discussed one effective design of the SPD manifold in fusion tasks. It cannot be overemphasized that the research space for multi-modal fusion tasks based on manifold learning remains vastly expansive. In the future, we will continue to explore new methods for solving classical computer vision tasks using other manifolds.

\section*{Availability of Data and Materials}
Information on access to the datasets supporting the conclusions of this article is included therein.

% Authors must disclose all relationships or interests that 
% could have direct or potential influence or impart bias on 
% the work: 
%
\section*{Competing Interests}
The authors declare that they have no conflict of interest.

\begin{acknowledgements}
This work was supported by the National Natural Science Foundation of China (62020106012, U1836218, 62106089, 62202205), the 111 Project of Ministry of Education of China (B12018), the Engineering and Physical Sciences Research Council (EPSRC) ( EP/V002856/1).
\end{acknowledgements}

% BibTeX users please use one of
\bibliographystyle{spbasic}      % basic style, author-year citations
%\bibliographystyle{spmpsci}      % mathematics and physical sciences
%\bibliographystyle{spphys}       % APS-like style for physics
%\bibliography{}   % name your BibTeX data base
\bibliography{ref.bib}

% % Non-BibTeX users please use
% \begin{thebibliography}{}
% %
% % and use \bibitem to create references. Consult the Instructions
% % for authors for reference list style.
% %
% \bibitem{RefJ}
% % Format for Journal Reference
% Author, Article title, Journal, Volume, page numbers (year)
% % Format for books
% \bibitem{RefB}
% Author, Book title, page numbers. Publisher, place (year)
% % etc
% \end{thebibliography}

\end{document}